\documentclass[lettersize,journal]{IEEEtran}
\usepackage{amsmath,amsfonts}
\usepackage{array}
\usepackage{textcomp}
\usepackage{stfloats}
\usepackage{url}
\usepackage{verbatim}
\usepackage{graphicx}
\usepackage{cite}
\usepackage{ulem}
\usepackage{subcaption}
\usepackage[colorlinks=true, urlcolor=blue, linkcolor=black]{hyperref}
\usepackage{cleveref}
\usepackage{booktabs}
\usepackage{xcolor}%
\definecolor{nblue}{rgb}{0.075,0.541,0.855}
\definecolor{mblue}{rgb}{0.075,0.541,0.855}
\definecolor{NavyBlue}{rgb}{0,0,0.5}
\usepackage{svg}
\usepackage{url}%
\usepackage[utf8]{inputenc}
\usepackage{multirow}
\usepackage{epstopdf}

\usepackage{scalerel}
\usepackage{tikz}
\usetikzlibrary{svg.path}

\definecolor{orcidlogocol}{HTML}{A6CE39}
\tikzset{
	orcidlogo/.pic={
		\fill[orcidlogocol] svg{M256,128c0,70.7-57.3,128-128,128C57.3,256,0,198.7,0,128C0,57.3,57.3,0,128,0C198.7,0,256,57.3,256,128z};
		\fill[white] svg{M86.3,186.2H70.9V79.1h15.4v48.4V186.2z}
		svg{M108.9,79.1h41.6c39.6,0,57,28.3,57,53.6c0,27.5-21.5,53.6-56.8,53.6h-41.8V79.1z M124.3,172.4h24.5c34.9,0,42.9-26.5,42.9-39.7c0-21.5-13.7-39.7-43.7-39.7h-23.7V172.4z}
		svg{M88.7,56.8c0,5.5-4.5,10.1-10.1,10.1c-5.6,0-10.1-4.6-10.1-10.1c0-5.6,4.5-10.1,10.1-10.1C84.2,46.7,88.7,51.3,88.7,56.8z};
	}
}

\newcommand\orcidicon[1]{\href{https://orcid.org/#1}{\mbox{\scalerel*{
				\begin{tikzpicture}[yscale=-1,transform shape]
				\pic{orcidlogo};
				\end{tikzpicture}
			}{|}}}}

\def\BibTeX{{\rm B\kern-.05em{\sc i\kern-.025em b}\kern-.08em
    T\kern-.1667em\lower.7ex\hbox{E}\kern-.125emX}}
\begin{document}
%
\title{Physically Consistent Image Augmentation for Deep Learning in Mueller Matrix Polarimetry}
\author{
    Christopher Hahne$^{\star}$$^{\textsuperscript{\orcidicon{0000-0003-2786-9905}}}$, 
    Omar Rodr\'{\i}guez-N\'u{\~n}ez$^{\textsuperscript{\orcidicon{0000-0003-0280-9519}}}$, 
    \'El\'ea Gros$^{\textsuperscript{\orcidicon{0000-0002-0911-2792}}}$, 
    Th\'eotim Lucas$^{\textsuperscript{\orcidicon{0000-0001-8509-4701}}}$,
    Ekkehard Hewer$^{\textsuperscript{\orcidicon{0000-0002-9128-0364}}}$, \\
    Tatiana Novikova$^{\textsuperscript{\orcidicon{0000-0002-9048-9158}}}$,
    Theoni Maragkou$^{\textsuperscript{\orcidicon{0000-0002-7774-760X}}}$,
    Philippe Schucht$^{\textsuperscript{\orcidicon{0000-0002-8025-3694}}}$,
    Richard McKinley$^{\textsuperscript{\orcidicon{0000-0001-8250-6117}}}$
    %
    \thanks{This work is funded by the Swiss National Science Foundation (SNSF) Sinergia Grant No. CRSII5\_205904. $^{\star}$Corresponding author email: \href{mailto:christopher.hahne@unibe.ch}{\textcolor{nblue}{christopher.hahne [att] unibe.ch}}
    }
    \thanks{C. Hahne is with the 
    Department of BioMedical Research (DBMR), University of Bern, 
    Bern, Switzerland.
    }
    \thanks{O. Rodr\'{\i}guez N\'u{\~n}ez and P. Schucht are with the Department of Neurosurgery, Inselspital, Bern, Switzerland.}
    \thanks{{\'E}. Gros and T. Maragkou are with the Institute of Tissue Medicine and Pathology, University of Bern, Bern, Switzerland.}
    \thanks{T. Lucas and T. Novikova are with the LPICM, CNRS, {\'E}cole polytechnique, Palaiseau, Paris, France}
    \thanks{E. Hewer is with the Institute of Pathology Lausanne University Hospital, University of Lausanne, Lausanne, Switzerland}
    \thanks{R. McKinley is with the 
    Support Center for Advanced Neuroimaging (SCAN), University Institute of Diagnostic and Interventional Neuroradiology, University of Bern, Inselspital,
    Bern, Switzerland.
    }
}

\markboth{Journal of \LaTeX\ Class Files,~Vol.~14, No.~8, August~2021}%
{Shell \MakeLowercase{\textit{et al.}}: A Sample Article Using IEEEtran.cls for IEEE Journals}


\maketitle

\begin{abstract}
Mueller matrix polarimetry captures essential information about polarized light interactions with a sample, presenting unique challenges for data augmentation in deep learning due to its distinct structure. %
While augmentations are an effective and affordable way to enhance dataset diversity and reduce overfitting, standard transformations like rotations and flips do not preserve the polarization properties in Mueller matrix images. %
To this end, we introduce a versatile simulation framework that applies physically consistent rotations and flips to Mueller matrices, tailored to maintain polarization fidelity. %
Our experimental results across multiple datasets reveal that conventional augmentations can lead to falsified results when applied to polarimetric data, underscoring the necessity of our physics-based approach.
In our experiments, we first compare our polarization-specific augmentations against real-world captures to validate their physical consistency. We then apply these augmentations in a semantic segmentation task, achieving substantial improvements in model generalization and performance. This study underscores the necessity of physics-informed data augmentation for polarimetric imaging in deep learning (DL), paving the way for broader adoption and more robust applications across diverse research in the field. In particular, our framework unlocks the potential of DL models for polarimetric datasets with limited sample sizes. %
Our code implementation is available at \texttt{{\color{mblue}\textmd{\href{https://github.com/hahnec/polar_augment}{github.com/hahnec/polar\_augment}}}}. %
\end{abstract}


\section{Introduction}
\label{sec:introduction}

\IEEEPARstart{I}{mage} data 
augmentation plays a crucial role in training deep neural networks by enhancing dataset diversity and mitigating the risk of overfitting~\cite{shorten2019survey}. Without sufficient data variety, larger deep learning models may memorize training examples rather than learning generalizable features, which results in high training accuracy but reduced performance on unseen data. This makes data augmentation essential for achieving robust, generalizable models~\cite{wang2017effectiveness}. %

%
In scenarios with limited training data, such as in polarimetric imaging, data augmentation becomes even more essential. Recent applications in robotics demonstrate how polarimetric imaging can improve perception of complex materials and scene geometry~\cite{taglione2024polarimetric}, highlighting its growing relevance beyond traditional domains. For example, medical datasets are often constrained by the difficulty of acquiring large, annotated samples, making robust augmentation strategies a cornerstone for improving DL model generalization and performance~\cite{wang2017effectiveness}.

Typical image augmentations involve pseudo-random modifications of image data including cropping~\cite{shorten2019survey}, intensity and color variations~\cite{hahne2021plenopticam}, additive noise or geometric transformations~\cite{wang2017effectiveness,chlap2021review,xu2023comprehensive}. 
\par

Geometric augmentation techniques, such as rotations and image flips, are extensively used in natural images but face limitations when applied to polarimetric images. This is because polarimetric imaging captures not only the spectral intensity but also the polarization state of scattered light, which is defined by the orientation of the electric field's oscillation plane~\cite{JRR_TN2023,
chipman2018polarized,Novikova2023}. As a result, standard color image rotation algorithms fail to account for changes in polarization properties. In polarimetric imaging, these properties are represented using the Mueller matrix, a 4$\times$4 real-valued transfer matrix that relates the Stokes vectors of the input and output light beams after interaction with a sample~\cite{Azzam2016,Goldstein,arteaga2022}. 
While the Mueller matrix captures full polarimetric information, conventional augmentations ignore its unique structure, highlighting the need for tailored transformation methods to prevent misleading results and support generalization.
Tailoring physics-informed data augmentation specific to imaging modalities has been explored in other fields such as ultrasound~\cite{tirindelli2021rethinking}, magnetic resonance imaging (MRI)\cite{pmlr-v139-fabian21a}, and light-fields~\cite{chao2024maskblur}. %
Existing augmentation techniques for polarimetric data are predominantly tailored to radar systems in astronomy and remote sensing, where physical modeling is mainly limited to rotation and mirror transformations~\cite{chen2013uniform,brown2014equivalence,aghababaei2023deep}. However, these methods fail to account for the unique structure of the Mueller matrix, making them unsuitable for optical polarimetric imaging applications. %
Previous studies on rotation-invariant parameters of the Mueller matrix and photon coordinate transformations in backscattering polarimetry provide a valuable groundwork for incorporating physical properties into augmentation methods~\cite{He2013, Li2018, Hao2024}. %
Building on this foundation, our work translates physical polarimetry models directly into deep learning-based augmentations, paving the way for more representative and effective training datasets, which is crucial for advancing optical polarimetric imaging applications that are constrained by data limitations. %
\par
Although prior research has acknowledged the challenges of augmenting optical polarimetric images~\cite{blanchon2020polarimetric,RUFFINO2022103495}, it lacks a rigorous mathematical framework for handling Mueller matrix transformations. Additionally, evaluations in existing work rely primarily on segmentation scores without quantitative comparisons to measured data, underscoring the need for validated augmentation methods that fully account for the unique structure of Mueller matrix images.
\par
Polarimetric imaging holds promise across various domains, but its potential is often hindered by the scarcity and limited variability of Mueller matrix datasets~\cite{gros2023effects,Giannantonio2023}. Existing augmentation methods typically neglect the physics of polarization, resulting in poor generalization of deep learning models. Our research addresses this gap by introducing a principled augmentation framework tailored to the structural and physical characteristics of Mueller matrices. This approach enhances model performance in tasks such as semantic segmentation~\cite{Krishna:2011,sampaio2023muller,hahne24muller}, denoising~\cite{Yang:22,moriconi2024denoising}, polarimetric demosaicking~\cite{Wen:2021,Pistellato:2022,Luo:2024}, and Mueller matrix recovery~\cite{chae2024machinelearning}, particularly in data-constrained environments where conventional methods fall short~\cite{gros2023effects,Giannantonio2023}. %
\par
To this end, we present a simulation framework that extends traditional spatial transformations to polarimetric imaging by systematic modeling of changes in the Mueller matrix elements. Our approach ensures that augmentations follow the physical laws of polarization, closely simulating how light would interact with materials and surfaces in real-world conditions.
Specifically, our findings reveal that rotating Mueller matrix images without modifying the polarization states results in erroneous polarization characteristics. By accounting for both spatial and polarization-specific variations, we can improve the generalization and accuracy of DL models in polarimetric imaging tasks. \par
The paper is structured as follows: the Methods section reviews standard spatial transformations and introduces isometric polarimetric mappings. The Results section presents key polarimetry features, metrics, and experimental validations of our framework. We then demonstrate the effectiveness of the proposed augmentations in training a segmentation network and conclude with a discussion.

\section{Methods}
Spatial transformations are fundamental for image augmentation in deep learning tasks, particularly effective for mitigating overfitting in data-limited scenarios. However, when applied to polarimetric imaging data, such transformations must account for the underlying physics of polarization measurement. In this work, we develop augmentation strategies tailored to Mueller matrix polarimetry by embedding physical constraints into spatial operations such as rotations and flips. \par
\begin{figure}[h]
    \centering
    \begin{minipage}[t]{0.49\linewidth}
        \centering
        \includegraphics[height=.8\linewidth]{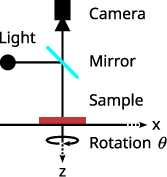}
        \subcaption{On-axis $(\alpha=0)$\label{fig:schematic:setup:a}}
    \end{minipage}
    \hfill
    \begin{minipage}[t]{0.49\linewidth}
        \centering
        \includegraphics[height=.8\linewidth]{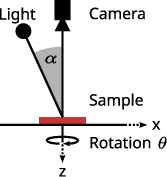}
        \subcaption{Off-axis $(\alpha\neq0)$\label{fig:schematic:setup:b}}
    \end{minipage}
    \caption{\textbf{Cross-sectional polarimeter setups}, showing collinear illumination and detection directions in (\subref{fig:schematic:setup:a}) and separated source and detector axes in (\subref{fig:schematic:setup:b}) spanning the tilt angle $\alpha$.}
    \label{fig:schematic:setup}
\end{figure}

%
%
Our approach assumes a quasi-collinear imaging configuration, where the illumination and detection axes are approximately aligned (see Fig.~\ref{fig:schematic:setup:a}). This assumption preserves the rotational symmetries of the Mueller matrix and ensures that transformed measurements remain physically interpretable. As illustrated in Fig.~\ref{fig:schematic:setup:b}, even modest off-axis tilts (nonzero angle $\alpha$ between source and detector) can introduce deviations from this symmetry and potentially lead to view-dependent polarization artifacts. 
For cases $\alpha\neq0$, we propose that image rotations about the detector's optical axis, $\theta \in \left[-4\pi/\alpha,+4\pi/\alpha\right]$, serve as a heuristic approximation range to ensure effective data augmentation results. %
\par
Nonetheless, for small angular separations, such distortions are typically negligible, and the proposed transformations remain valid approximations. This is analogous to how specular reflections in conventional imaging can violate intensity constancy without compromising the overall utility of geometric augmentations. These practical considerations define the boundary conditions under which our augmentation framework is expected to generalize reliably. \par%
We present our Mueller matrix image augmentations in two parts: the first revisits spatial isometric mappings, such as rotations and flips, to establish a foundation for integrating their polarimetry-specific consequences. 

\subsection{Spatial transformations}

A spatial isometric mapping of an image with coordinates $\mathbf{x}_i = (x_i, y_i)^\top $ is given by

\begin{align}
\mathbf{x}_i' = \mathbf{T}_s \left(\mathbf{x}_i-\mathbf{c}\right)+\mathbf{c} \, ,
\label{eq:spatial}
\end{align}
where $\mathbf{x}_i' = (x_i', y_i')^\top$ are the transformed coordinates, $\mathbf{c} = (x_c, y_c)^\top$ is the image center, and $\mathbf{T}_s\in \text{O}(2)$ is an isometric spatial transformation. Here, $i\in\{0, 1, \dots, H\times W-1\}$ is a pixel index, where $H$ and $W$ denote the image's vertical and horizontal dimensions. Having defined the general spatial transformation matrix $\mathbf{T}_s$, we plan to substitute it with any isometric matrix (e.g., rotation matrix) hereafter. \par

The rotation of a two-dimensional (2-D) coordinate vector $\mathbf{x}_i$ by an angle $\theta \in \left[0,2\pi\right)$ is described by the special orthogonal group $\text{SO}(2)$ such that $R_s(\theta) \in \text{SO}(2)$ is:

\begin{align}
\mathbf{R}_s=R_s(\theta) = 
\begin{bmatrix}
\cos\theta & -\sin\theta \\
\sin\theta & \cos\theta
\end{bmatrix}.
\label{eq:spatial_rota}
\end{align}

This matrix rotates \( \mathbf{x}_i \) counter-clockwise and preserves the Euclidean norm, ensuring that the distance between points remains unchanged under transformation, a fundamental property of \(\text{SO}(2)\) group. \par

For the spatial image flipping, we have
\begin{align}
\mathbf{H}_s = 
\begin{bmatrix}
1 & 0 \\
0 & -1
\end{bmatrix} , \,\,\,
\mathbf{V}_s = 
\begin{bmatrix}
-1 & 0 \\
0 & 1
\end{bmatrix} , \,\,\,
\mathbf{F}_s = 
\begin{bmatrix}
-1 & 0 \\
0 & -1
\end{bmatrix} ,
\label{eq:spatial:flip}
\end{align}
where $\mathbf{H}_s$ is the horizontal and $\mathbf{V}_s$ the vertical flip matrix, which yield $\mathbf{F}_s = \mathbf{H}_s\mathbf{V}_s$ when being combined.

A spatial image rotation moves areas outside the field-of-view into the sampling grid. In grayscale or color images, these missing areas are typically handled by padding strategies that assign default values to the affected pixels. In the case of Mueller matrix images, however, we are dealing with intensity data used for algebra operations, which demands a different treatment. For instance, computing the inverse of a $4\times 4$ matrix filled with zeros is an undefined operation because such matrix is singular. In DL, this would propagate to an undefined value for the training loss. To overcome this problem, we propose to either fill the 16 Mueller matrix channels with a flattened identity matrix $\mathbf{I}_4=\operatorname{diag}(1, 1, 1, 1)$ or mirror the affected pixels at the image boundary.

After applying a transformation $\mathbf{T}_s$, the coordinates \( \mathbf{x}_i' \) typically do not align with the discrete grid points in the original spatial domain. Therefore, interpolation is required to estimate the values at the new coordinates. Common interpolation methods include nearest-neighbor, bilinear, and bicubic interpolation, with bilinear interpolation often serving as a compromise between computational efficiency and image quality. The interpolation ensures that the rotated image maintains visual fidelity, minimizing artifacts introduced by the coordinate transformation.

\begin{figure*}[!b]
	\centering
	\begin{minipage}{0.19\textwidth}
		\centering
		\includegraphics[width=\textwidth]{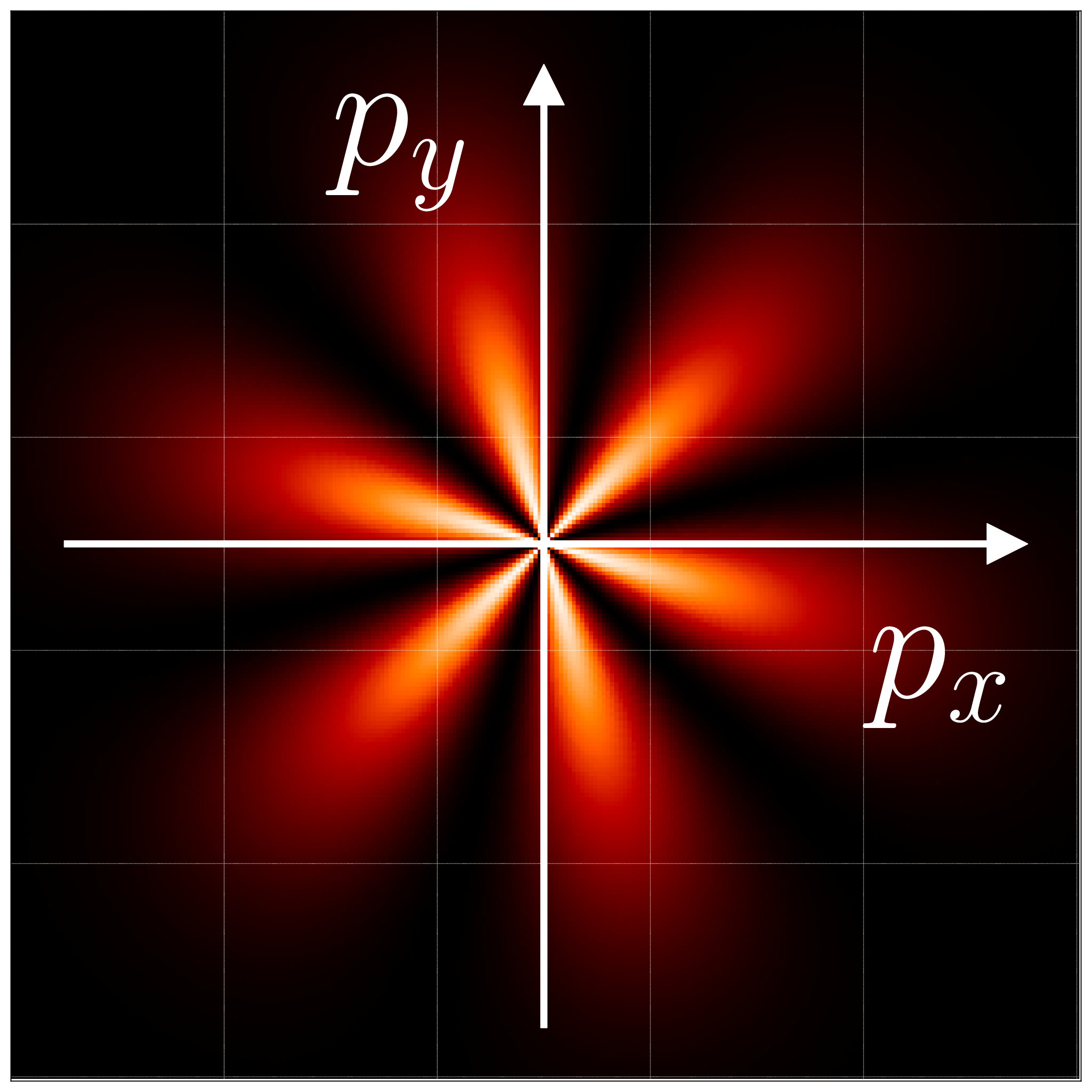}
		\subcaption{Reference\label{subfig:a}}
	\end{minipage}
	\hfill
	\begin{minipage}{0.19\textwidth}
		\centering
		\includegraphics[width=\textwidth]{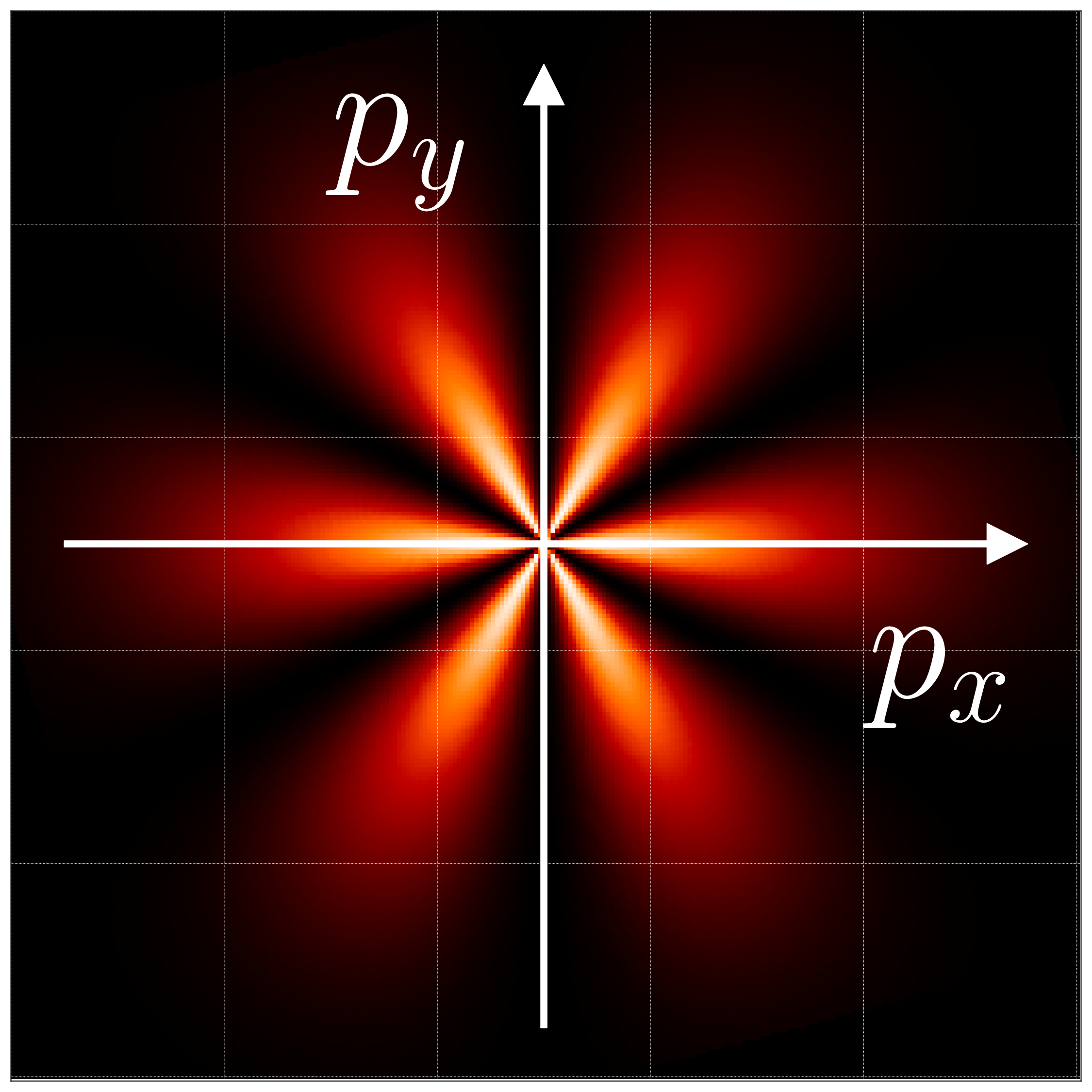}
		\subcaption{Rot. $\theta=15$\textdegree\label{subfig:b}}
	\end{minipage}
	\hfill
	\begin{minipage}{0.19\textwidth}
		\centering
		\includegraphics[width=\textwidth]{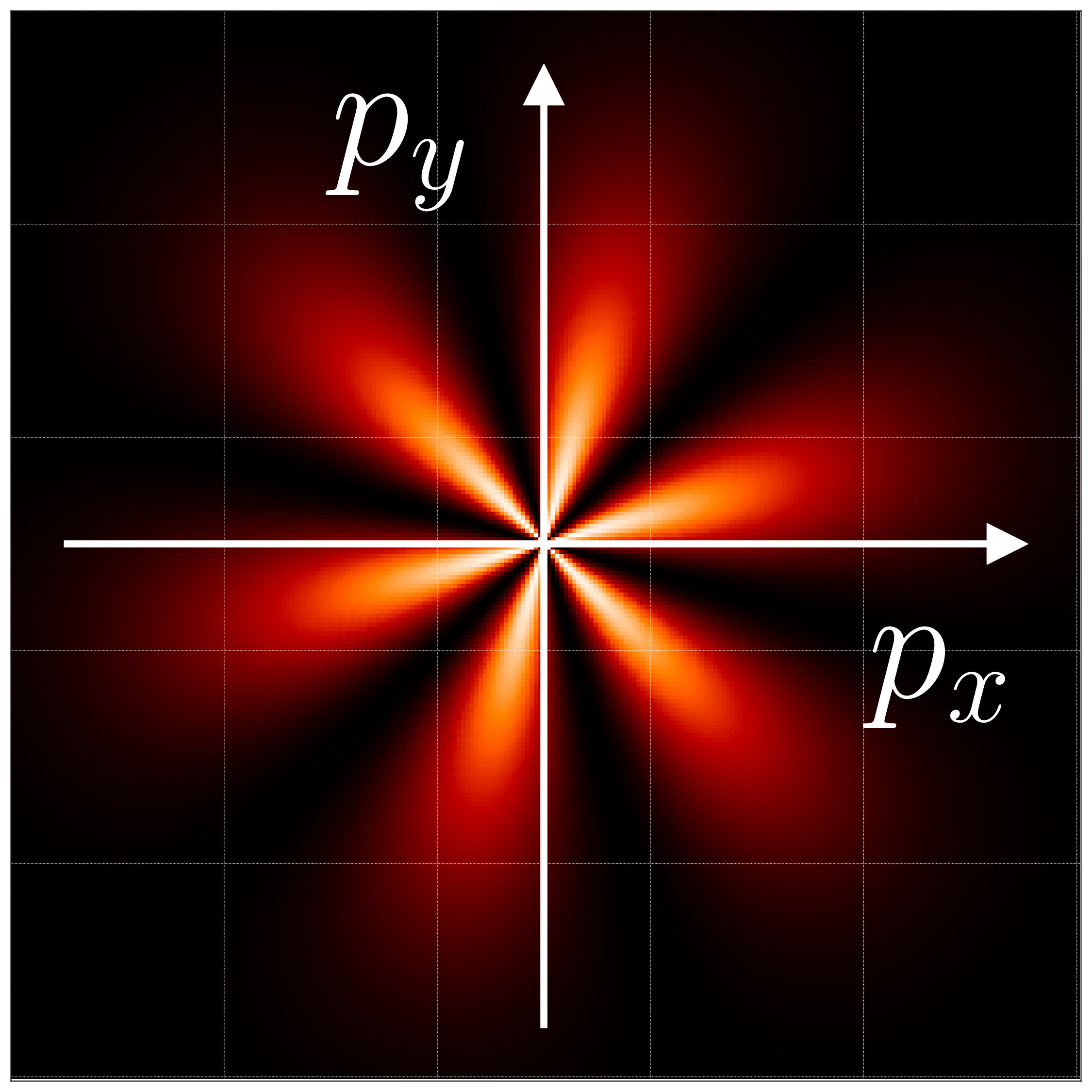}
		\subcaption{Horizontal flip\label{subfig:c}}
	\end{minipage}
	\hfill
	\begin{minipage}{0.19\textwidth}
		\centering
		\includegraphics[width=\textwidth]{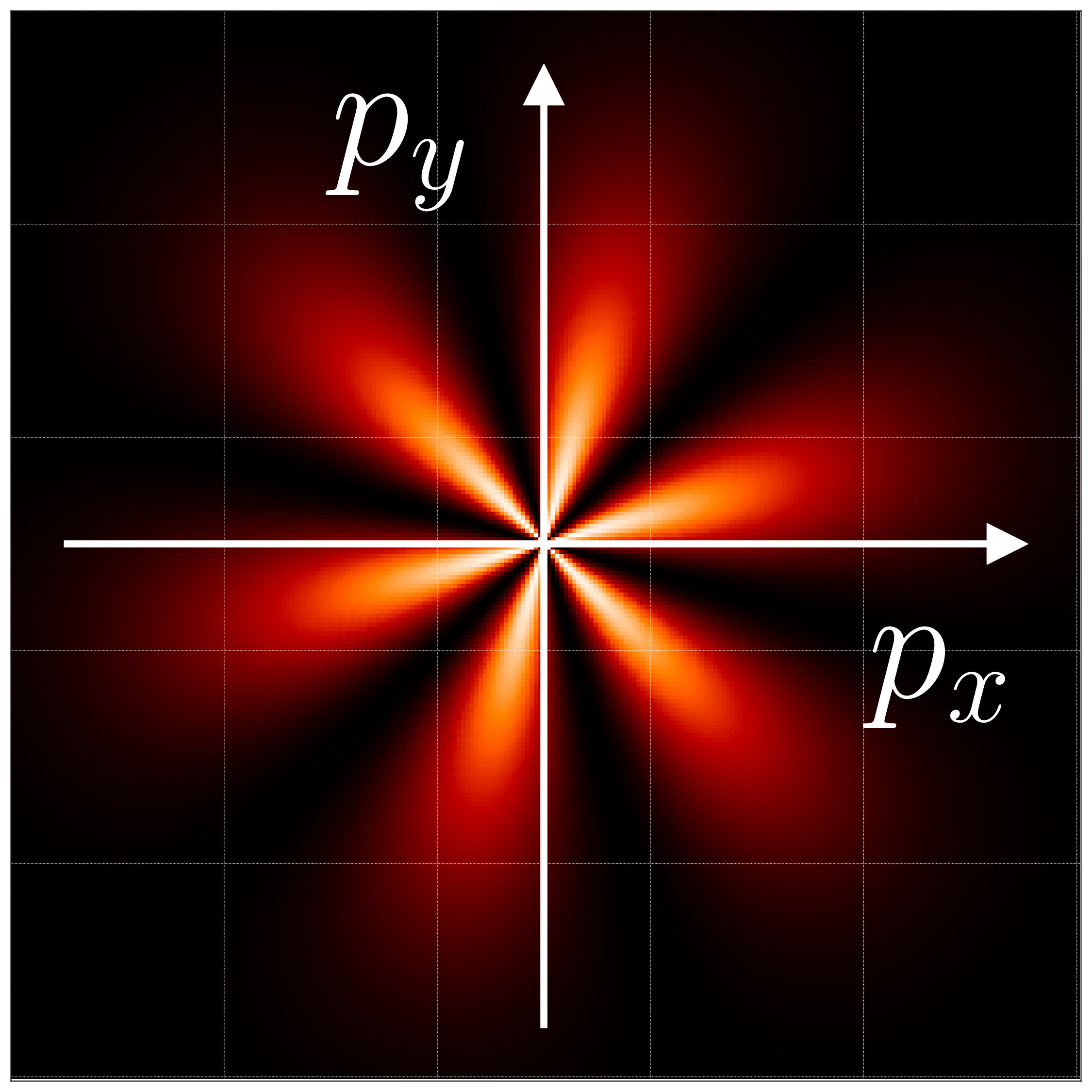}
		\subcaption{Vertical flip\label{subfig:d}}
	\end{minipage}
	\hfill
	\begin{minipage}{0.19\textwidth}
		\centering
		\includegraphics[width=\textwidth]{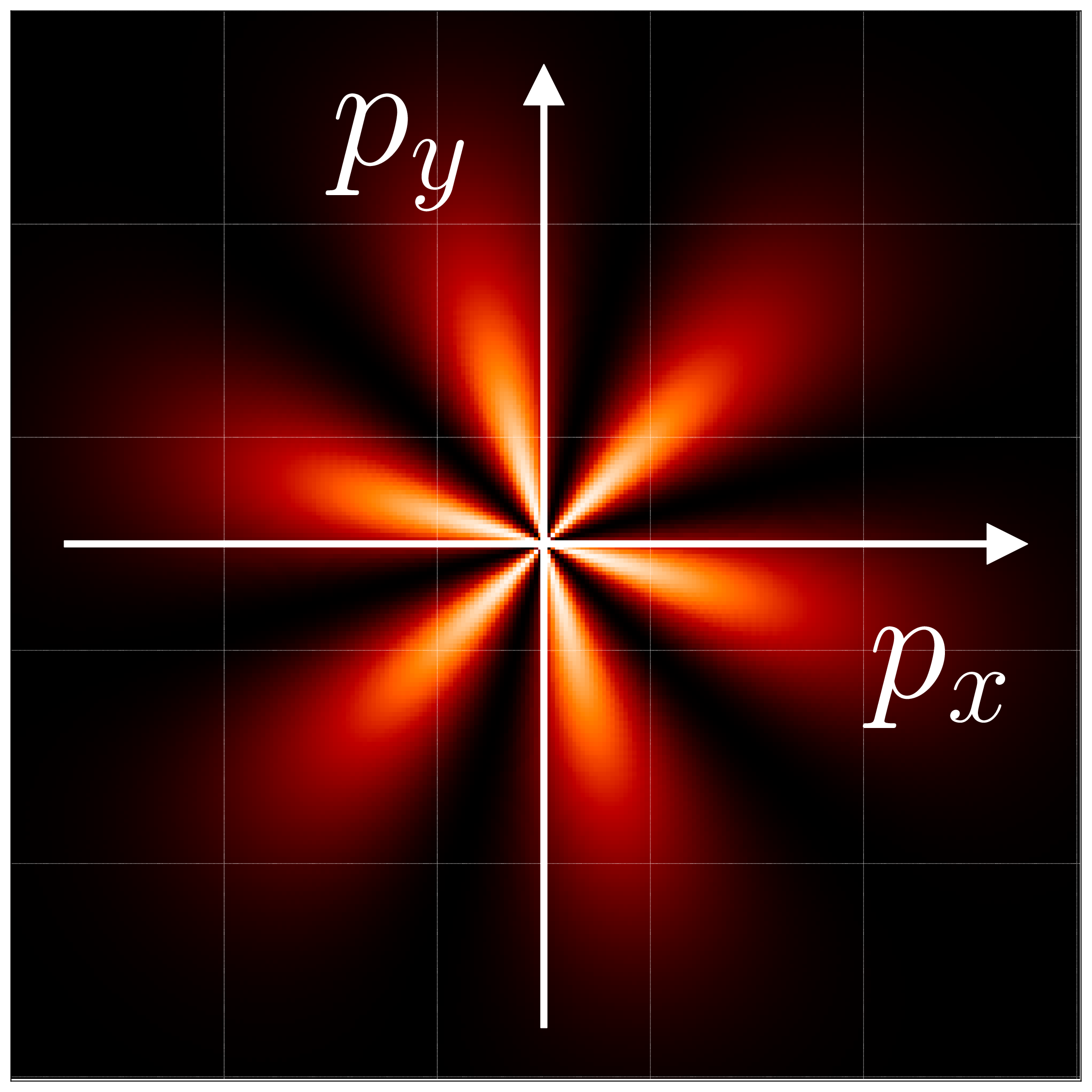}
		\subcaption{Both flips\label{subfig:e}}
	\end{minipage}
	\caption{\textbf{Physical model for polarimetry states transformation.} The images show intensities modeled by sinusoidal distribution in the angular domain. The axes $p_x$ and $p_y$ are perpendicular to the direction of ray propagation. The orientation of the polarization plane for linearly polarized states 
		is identical for (\subref{subfig:c}) the horizontal and (\subref{subfig:d}) the vertical flip as well as for the input (\subref{subfig:a}) and (\subref{subfig:e}) a combined horizontal and vertical flip.\label{fig:states}}
\end{figure*}
\subsection{Polarimetric transformations}

When $\mathbf{M}_i\in\mathbb{R}^{4\times4}$ denotes the captured Mueller matrix at image pixel $i$, the polarimetric transformation is given by
\begin{align}
    \mathbf{M}_i' &= \mathbf{T}_p \,\mathbf{M}_i\,\mathbf{T}_p^{-1}\label{eq:mmpolar}
    ,
\end{align}
where $\mathbf{T}_p\in \text{O}(4)$ is an orthogonal matrix. For notational clarity, the $(u,v)$-th scalar entry of any bold matrix is written as $M_i^{(u,v)}$. %
We assume a rigid transformation of the entire instrument about its optical axis leaving the measured intensity~$\mathbf B_i$ invariant. %

In a deep learning scenario, we wish to augment polarimetry images and equally apply the transform to the corresponding ground truth (GT) label. However, GT labels are unavailable in cases where the Mueller matrix formation is part of a model process. This motivates us to embed the augmentation transforms into the calibration data before Mueller matrix computation as it is more generic, enabling augmentation within the data loading process in PyTorch. 
The Mueller matrix $\mathbf{M}_i$ is obtained as follows,
\begin{align}
    \mathbf{M}_i &= \mathbf{A}_i^{-1} \mathbf{B}_i^{} \mathbf{W}_i^{-1} \label{eq:mm}
\end{align}
where $\mathbf{B}_i\in\mathbb{R}^{4\times 4}$ contains measured intensities per pixel $i$ and $\mathbf{A}_i\in\mathbb{R}^{4\times 4}$ and $\mathbf{W}_i\in\mathbb{R}^{4\times 4}$ are constructed using four linear independent Stokes vectors as their columns and rows, respectively. The choice of optimal polarization states (i.e., corresponding Stokes vectors) for polarization modulation and analysis is guided by minimization and equal distribution of experimental noise across all coefficients of the Mueller matrix. Therefore, both $\mathbf{A}_i$ and $\mathbf{W}_i$ should be well-conditioned to ensure reliable inversion. As shown in prior work~\cite{Gribble:13}, an optimal design of a complete Mueller polarimeter requires the four Stokes vectors to form a regular tetrahedron on the Poincaré sphere~\cite{Goldstein}. 
In practice, the experimentally obtained matrices $\mathbf{A}_i$ and $\mathbf{W}_i$ often deviate from this ideal configuration due to imperfections of the system optical components and mechanical alignment tolerances. The calibration process involves determination of these experimental matrices to characterize the system’s polarization response. We calibrate our Mueller polarimeter using the Eigenvalue Calibration Method (ECM)~\cite{Compain1999}, which enables accurate system characterization without requiring precise optical modeling or mechanical alignment. In ECM, $\mathbf{W}_i$ and $\mathbf{A}_i$ are estimated from a minimal set of four reference samples, capturing both experimental imperfections and wavelength-dependent behavior of optical components. Thus, ECM enables unambiguous retrieval of polarimetric properties for both the instrument and the sample. \par

When plugging \eqref{eq:mm} into \eqref{eq:mmpolar}, we obtain
\begin{align}
    \mathbf{M}_i' &= \mathbf{T}_p \, \mathbf{A}_i^{-1} \mathbf{B}_i^{} \mathbf{W}_i^{-1} \, \mathbf{T}_p^{-1} \label{eq:mmext}
    .
\end{align}
By expanding the products of the transform and calibration matrices, we get
\begin{align}
    \mathbf{A}_i' = \left(\mathbf{T}_p \mathbf{A}_i^{-1}\right)^{-1} \quad \text{and} \quad 
    \mathbf{W}_i' = \left(\mathbf{W}_i^{-1} \mathbf{T}_p^{-1}\right)^{-1} \label{eq:awext}
    ,
\end{align}
which allows us to replace the original calibration matrices in \eqref{eq:mm} during data loading. To avoid calculating the inverse multiple times, we rearrange \eqref{eq:awext} to
\begin{align}
    \mathbf{A}_i' = \mathbf{A}_i \mathbf{T}_p^{-1} \quad \text{and} \quad 
    \mathbf{W}_i' = \mathbf{T}_p \mathbf{W}_i \label{eq:awext:rearrange}
    ,
\end{align}
for computational efficiency. From this, we obtain the transformed Mueller matrix by adapted matrices
\begin{align}
    \mathbf{M}_i' &= \left(\mathbf{A}_i'\right)^{-1} \mathbf{B}_i^{} \left(\mathbf{W}_i'\right)^{-1} \label{eq:mmext_}
    ,
\end{align}
which enables polarimetry augmentation by only modifying the calibration data. It is worth mentioning that the polarimetry augmentation is applied directly to the Stokes vectors, which are used to construct the matrices $\mathbf{A}_i$ and $\mathbf{W}_i$. After defining the general polarimetric transformation matrix $\mathbf{T}_p$, we can now substitute it with the following isometric matrices for a rotation or image flip within the polarimetric domain.

Given the angle $\theta \in \left[0,2\pi\right)$, we let $R_p(\theta)$ represent the rotational change of basis matrix for Stokes vectors and Mueller matrices, which is defined as
\begin{align}
\mathbf{R}_p = R_p(\theta) = 
\begin{bmatrix}
1 & 0 & 0 & 0 \\
0 & \cos 2\theta & -\sin 2\theta & 0 \\
0 & \sin 2\theta & \cos 2\theta & 0 \\
0 & 0 & 0 & 1
\end{bmatrix}
\label{eq:rota}
,
\end{align}
for the counter-clockwise direction~\cite{brown2014equivalence,Goldstein}.
When plugging Eq.~\eqref{eq:rota} into $\mathbf{T}_p$ in Eqs.~\eqref{eq:mmpolar} or \eqref{eq:awext}, the relation $R_p(\theta)^{-1}=R_p(-\theta)$ simplifies the inverse computation given the orthogonality property of a rotation matrix. \par

For the polarimetric flipping, we aim to mimic a perfect mirror at normal incidence of light as a transformation matrix assuming it behaves like an ideal linear reflector without the introduction of significant depolarization or retardance effects. The key effect of a mirror on polarization is that it flips the handedness of circular polarization and the orientation of linear ($\pm$45\textdegree) polarization planes, but the intensity of linear horizontally and vertically polarized light remain unaffected
~\cite{brown2014equivalence}. %
For example, if the light is horizontally polarized, its electric field oscillates in the horizontal plane. Upon reflection, the direction of propagation changes, but the oscillation remains in the horizontal plane. Similarly, if the light is vertically polarized, the same logic applies; the electric field oscillates in the vertical plane both before and after reflection. \par
Following the approach of \cite{Li2018}, we base our transformation on the mirror symmetry of the sample, employing the transformation $H_p(\theta)$:
\begin{align}
\mathbf{H}_p=H_p(\theta) = 
\begin{bmatrix}
1 & 0 & 0 & 0 \\
0 & \cos 4\theta & \sin 4\theta & 0 \\
0 & \sin 4\theta & -\cos 4\theta & 0 \\
0 & 0 & 0 & -1
\end{bmatrix} \, .
\label{eq:polar:flip:general}
\end{align}

In the case of $\theta \in \{k \frac{\pi}{2} \mid k \in \mathbb{Z} \}$, this matrix simplifies to:
\begin{align}
\mathbf{H}_k = 
H_p\left(k\frac{\pi}{2}\right) = 
\begin{bmatrix}
1 & 0 & 0 & 0 \\
0 & 1 & 0 & 0 \\
0 & 0 & -1 & 0 \\
0 & 0 & 0 & -1
\end{bmatrix} , \, 
%
\label{eq:polar:flip}
\end{align}
%
which directly describes a horizontal or vertical flipping transformation, respectively. %
Figure~\ref{fig:states} illustrates these relationships for an intuitive comprehension.
%
%
Note that the product of the horizontal and vertical reflection matrices in our framework is equivalent to a $180^{\circ}$ rotation, i.e., $\mathbf{H}_k \mathbf{H}_k = R_p(\pi)$. Since both $\mathbf{H}^2_k$ and $R_p(\pi)$ equal the $4 \times 4$ identity matrix, this confirms the internal consistency of our transformation algebra.
We implement a combined flip as well as the horizontal and vertical flips in conjunction with the spatial transforms from Eq.~\eqref{eq:spatial:flip}.
\begin{figure*}[!b]
    \centering
    \begin{minipage}[b]{0.06\textwidth}
        \centering
        \raisebox{5.75\height}{30\textdegree}
    \end{minipage}%
    \hfill
    \begin{minipage}[b]{0.2\textwidth}
        \centering
        \includegraphics[width=\textwidth]{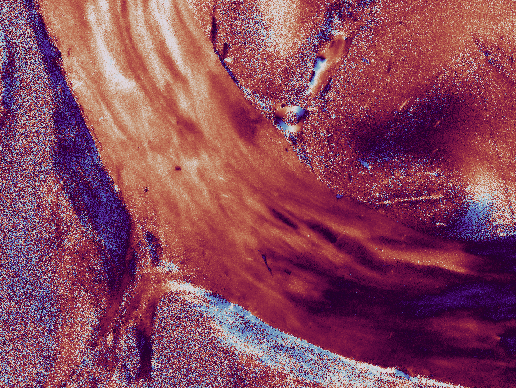}
    \end{minipage}%
    \hfill
    \begin{minipage}[b]{0.2\textwidth}
        \centering
        \includegraphics[width=\textwidth]{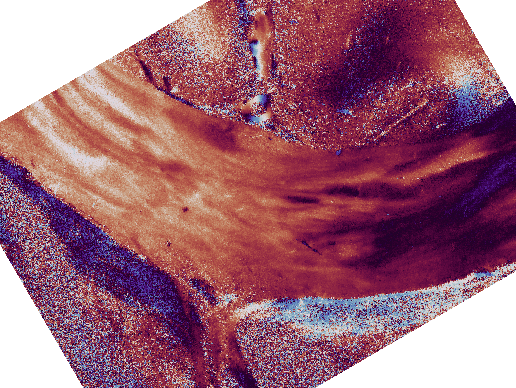}
    \end{minipage}%
    \hfill
    \begin{minipage}[b]{0.2\textwidth}
        \centering
        \includegraphics[width=\textwidth]{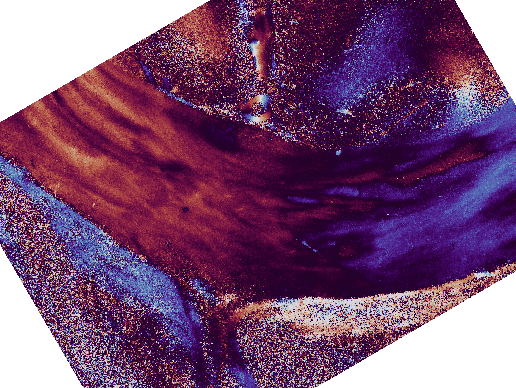}
    \end{minipage}%
    \hfill
    \begin{minipage}[b]{0.2\textwidth}
        \centering
        \includegraphics[width=\textwidth]{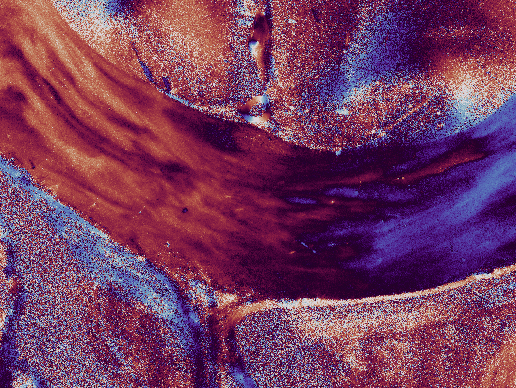}
    \end{minipage} \\[1ex]

    \begin{minipage}[b]{0.06\textwidth}
        \centering
        \raisebox{5.75\height}{60\textdegree}
    \end{minipage}%
    \hfill
    \begin{minipage}[b]{0.2\textwidth}
        \centering
        \includegraphics[width=\textwidth]{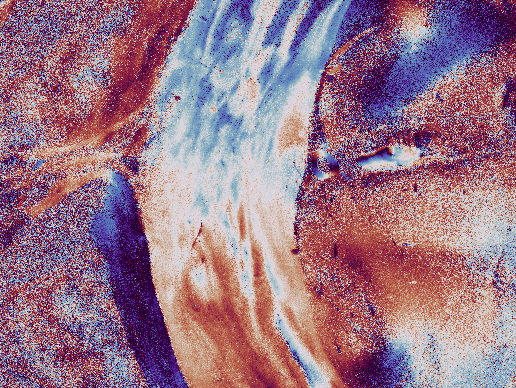}
    \end{minipage}%
    \hfill
    \begin{minipage}[b]{0.2\textwidth}
        \centering
        \includegraphics[width=\textwidth]{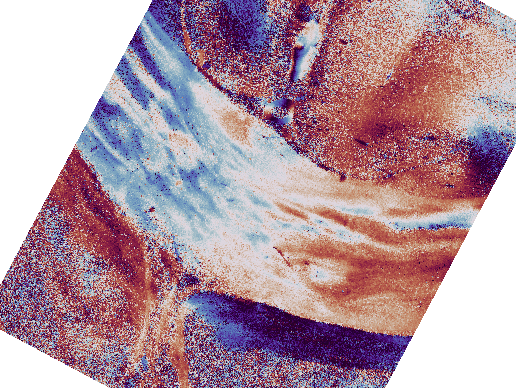}
    \end{minipage}%
    \hfill
    \begin{minipage}[b]{0.2\textwidth}
        \centering
        \includegraphics[width=\textwidth]{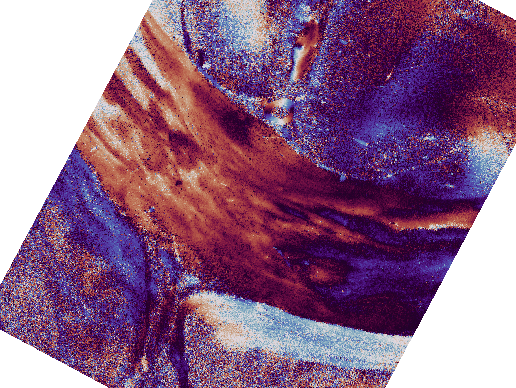}
    \end{minipage}%
    \hfill
    \begin{minipage}[b]{0.2\textwidth}
        \centering
        \includegraphics[width=\textwidth]{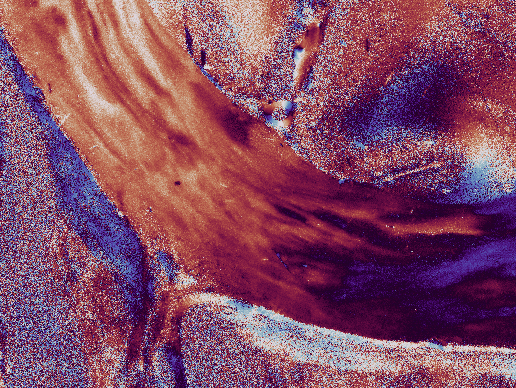}
    \end{minipage} \\[1ex]

    \begin{minipage}[b]{0.06\textwidth}
        \centering
        \raisebox{5.75\height}{80\textdegree}
    \end{minipage}%
    \hfill
    \begin{minipage}[b]{0.2\textwidth}
        \centering
        \includegraphics[width=\textwidth]{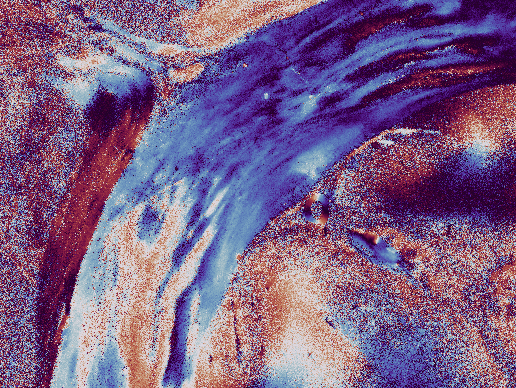}
    \end{minipage}%
    \hfill
    \begin{minipage}[b]{0.2\textwidth}
        \centering
        \includegraphics[width=\textwidth]{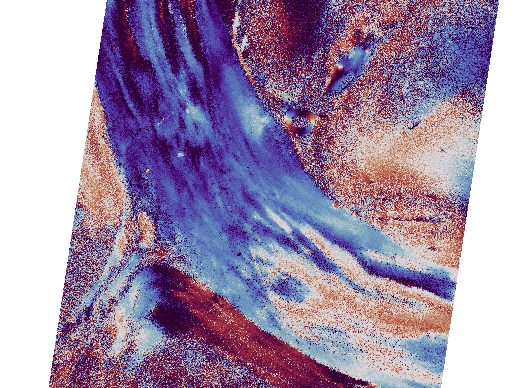}
    \end{minipage}%
    \hfill
    \begin{minipage}[b]{0.2\textwidth}
        \centering
        \includegraphics[width=\textwidth]{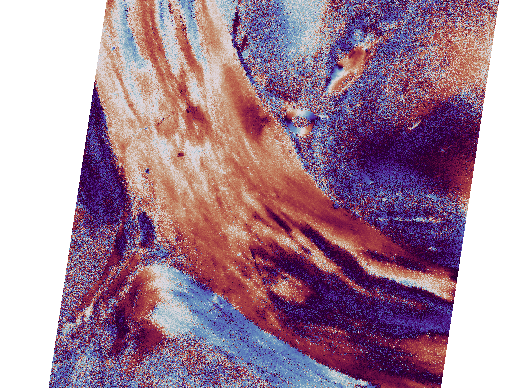}
    \end{minipage}%
    \hfill
    \begin{minipage}[b]{0.2\textwidth}
        \centering
        \includegraphics[width=\textwidth]{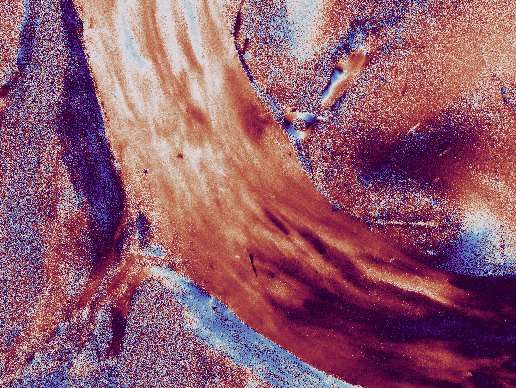}
    \end{minipage} \\[1ex]

    \begin{minipage}[b]{0.06\textwidth}
        \centering
        \raisebox{5.75\height}{90\textdegree}
    \end{minipage}%
    \hfill
    \begin{minipage}[b]{0.2\textwidth}
        \centering
        \includegraphics[width=\textwidth]{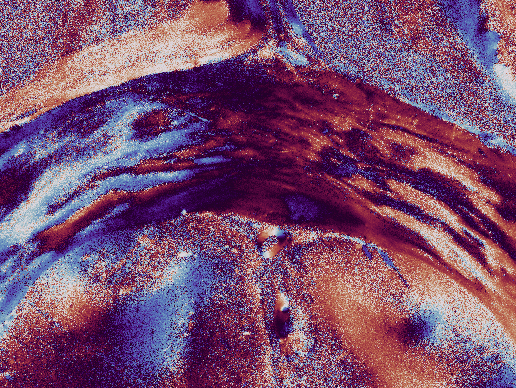}
    \end{minipage}%
    \hfill
    \begin{minipage}[b]{0.2\textwidth}
        \centering
        \includegraphics[width=\textwidth]{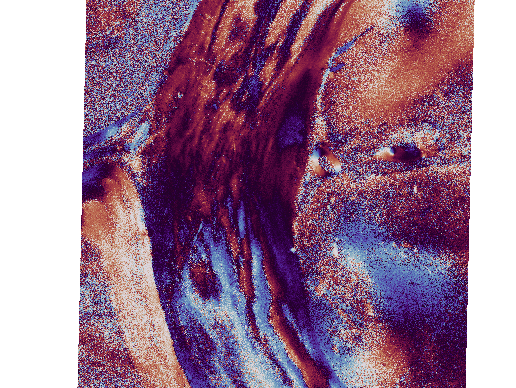}
    \end{minipage}%
    \hfill
    \begin{minipage}[b]{0.2\textwidth}
        \centering
        \includegraphics[width=\textwidth]{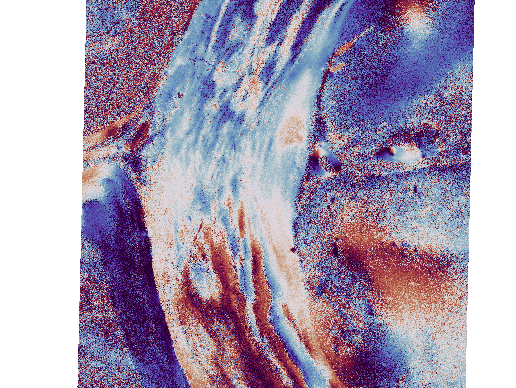}
    \end{minipage}%
    \hfill
    \begin{minipage}[b]{0.2\textwidth}
        \centering
        \includegraphics[width=\textwidth]{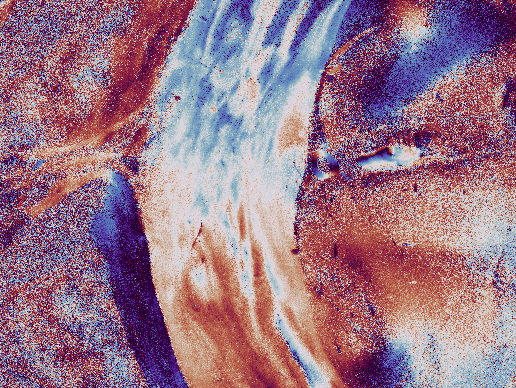}
    \end{minipage} \\[1ex]

    \begin{minipage}[b]{0.06\textwidth}
        \centering
        \raisebox{2.8\height}{\parbox{\textwidth}{Flip/\\180\textdegree}}
        \subcaption*{$\theta$}
    \end{minipage}%
    \hfill
    \begin{minipage}[b]{0.2\textwidth}
        \centering
        \includegraphics[width=\textwidth]{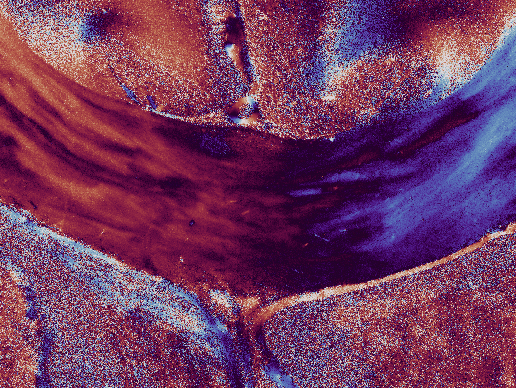}
        \subcaption*{Input}
    \end{minipage}%
    \hfill
    \begin{minipage}[b]{0.2\textwidth}
        \centering
        \includegraphics[width=\textwidth]{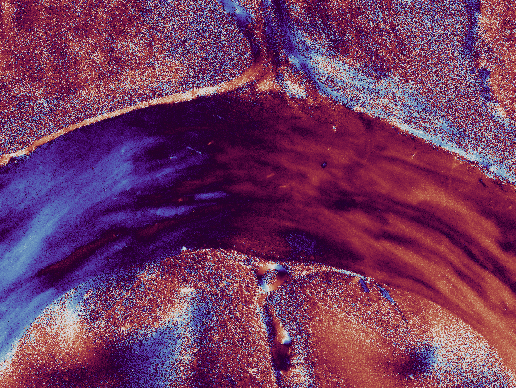}
        \subcaption*{After Eq.~\eqref{eq:spatial}}
    \end{minipage}%
    \hfill
    \begin{minipage}[b]{0.2\textwidth}
        \centering
        \includegraphics[width=\textwidth]{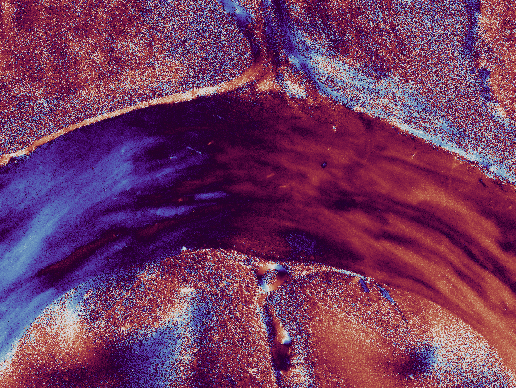}
        \subcaption*{After~Eqs.~\eqref{eq:spatial}~\&~\eqref{eq:mmpolar}}
    \end{minipage}%
    \hfill
    \begin{minipage}[b]{0.2\textwidth}
        \centering
        \includegraphics[width=\textwidth]{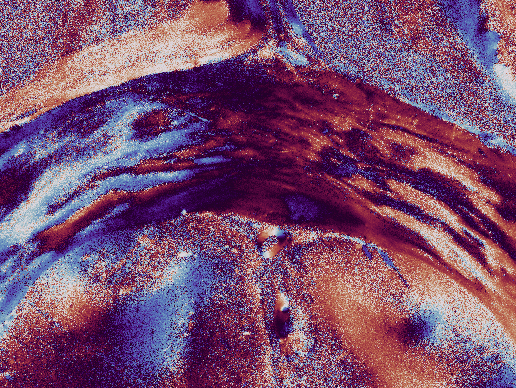}
        \subcaption*{Ground truth}
    \end{minipage} \\[1ex]
    \vspace{.05cm}
    \includegraphics[width=.8\linewidth]{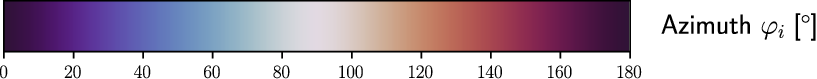}
    \vspace{.1cm}
    \caption{%
    \textbf{Image transformation results} that show the effectiveness of polarimetric rotation and flipping for azimuth images of formalin-fixed thick brain tissue specimen. The first column shows experimental images, and the last column presents images measured from a physically rotated sample at the indicated angles. Each row corresponds to a different rotation angle of a sample, with transformations applied across the columns. %
    }
    \label{fig:grid_of_images}
\end{figure*}

\section{Results}
We present a two-fold evaluation approach to validate our computational framework. First, we conduct numerical and qualitative comparisons between simulated transformations and an image-based dataset comprising 18 acquisitions from a single specimen, incrementally rotated in 10-degree steps. This dataset serves to validate our framework against real-world rotations, despite lacking annotations and being insufficient for training. To complement this, we utilize an additional annotated dataset for semantic segmentation training in brain-surgical tissue, allowing us to evaluate our proposed data augmentation techniques in a practical context.
\subsection{Image-based comparison}
For the validation of the proposed rotation and flip, we acquire 18 frames of tissue specimen with an incremental rotation of 10 degree step size. %
To extract human-interpretable images from polarimetric acquisitions, we decompose the transformed Mueller matrices $\mathbf{M}'_i\in\mathbb{R}^{4\times 4}$ according to Lu and Chipman~\cite{lu1996interpretation},
\begin{align}
    \mathbf{M}'_i = \boldsymbol{\Delta}_i \mathbf{R}_i \mathbf{D}_i \, ,
\end{align}
which is accomplished by the singular value decomposition. Here, $\boldsymbol{\Delta}_i$ denotes the depolarization matrix, $\mathbf{R}_i$ the retardance matrix, and $\mathbf{D}_i$ the diattenuation matrix, each representing physically meaningful properties of the sample at pixel $i$. %
In particular, we choose the per-pixel azimuth angle of the optical axis of a linear retarder from the Lu–Chipman polar decomposition to verify correct polarimetric rotation. The motivation comes from azimuth angles highlighting the orientation of brain fiber tracts in white matter tissue during neurosurgery~\cite{Rodriguez-Nunez:21,Felger:23,Novikova2023:2}.

In particular, we choose the per-pixel azimuth angle of the optical axis of a linear retarder from Lu-Chipman polar decomposition to verify correct polarimetric rotation. The motivation comes from azimuth angles highlighting the orientation of brain fiber tracts in white matter tissue during neurosurgery~\cite{Rodriguez-Nunez:21,Felger:23,Novikova2023:2}. %
The azimuth $\varphi_i \in \mathbb{R} \cap [0, \pi)$ is computed from a ratio of scalar-valued elements of the linear retardance matrix $\mathbf{R}_i$. We denote the $(u,v)$-th entry of this matrix as $R_i^{(u,v)}$, such that
\begin{align}
    \varphi_i = \frac{1}{2}\tan^{-1} \left(\frac{R_i^{(2,4)}}{R_i^{(4,3)}}\right), \label{eq:azimuth}
\end{align}
where $(u,v)$ are the row and column indices, respectively.
Figure~\ref{fig:grid_of_images} shows azimuth $\varphi_i$ maps at different rotation angles before and after applying transformations. The first column displays experimental images, while the last column presents images measured from a sample physically rotated by the indicated angles. Each row corresponds to the same sample, with the first image representing its initial position and subsequent columns illustrating the transformed versions. \par
For the numerical analysis, we compute the mean absolute error (MAE) taking into account that the azimuth is cyclic in the range $[0,\pi)$:
\begin{align}
    \text{MAE}_\varphi = \frac{1}{N} \sum^N_i \min\left(\left\vert\varphi'_i-\varphi^{(g)}_i\right\vert, \left\vert\varphi_i-\varphi^{(g)}_i\right\vert\right), 
\end{align}
where $\varphi'_i=\pi-\varphi_i$ and $\varphi^{(g)}_i$ denotes the ground truth, which is obtained by rotating the specimen in the real space during the experimental measurements. %
For benchmarking, we adapt the augmentation strategy from Blanchon~\textit{et al.}~\cite{blanchon2020polarimetric}, who apply spatial image rotation and compensate for physical consistency by adjusting the angle of polarization (AoP) accordingly. Specifically, we extend their method by applying a post-hoc angular correction, which we term $\theta$-Offset, directly to the extracted azimuth map $\varphi_i$ after Lu–Chipman decomposition. %
The errors are presented in Table~\ref{tab:error}. \par
%

\begin{table}[htbp]
\centering
\caption{Numerical deviations for azimuth $\varphi_i$.}
\begin{tabular}{l c c c}
\toprule
 & \textbf{Spatial-only} & $\theta$-\textbf{Offset}~\cite{blanchon2020polarimetric} & \textbf{Proposed} \\
\midrule
$\downarrow\,\text{MAE}_\varphi$ [deg] & 27.167 & 13.343 & 12.977 \\ 
\bottomrule\bottomrule
\end{tabular}
\label{tab:error}
\end{table}
For the comparison in Table~\ref{tab:error}, we mask regions outside the field-of-view and noisy background areas by thresholding the linear retardance $\delta_i \in \mathbb{R}$ at the 75\textsuperscript{th} percentile, using this as a heuristic criterion. The linear retardance $\delta_i$ is computed as: %
\begin{align}
    \delta_i&=\arccos\left(\sqrt{P_i}-1\right), \,\, \text{where} \\ 
    P_i&=\left(R_i^{(2,2)}+R_i^{(3,3)}\right)^2+\left(R_i^{(3,2)}-R_i^{(2,3)}\right)^2 \label{eq:retardance}
\end{align} following the definition by Chipman \textit{et al.}~\cite{chipman2018polarized}. 
\par
%
Table~\ref{tab:error} shows that our proposed method yields the lowest MAE in azimuth computation, indicating the closest alignment with experimental ground truth. In contrast, spatial-only transformations produce substantially larger errors, as they disregard the polarization-specific structure of the data. While the $\theta$-Offset method~\cite{blanchon2020polarimetric} performs comparably to our approach in terms of MAE, it relies on a simplified adjustment that is not applicable to the Mueller matrix and lacks a rigorous physical foundation.
The deviations observed in Table~\ref{tab:error} potentially arise from slight geometric misalignments between the illumination and detection axes relative to the specimen surface. Such off-axis configurations can break the quasi-collinear assumption underlying our augmentation model, disrupting the rotational symmetry of the polarimetric transform. This may introduce view-dependent polarization effects, particularly in regions with anisotropic scattering or surface curvature, which could contribute to the residual azimuth errors observed.
Additionally, noise in the measurements, particularly in the areas surrounding the tissue, contributes to these deviations. While this relationship is also observed in Fig.~\ref{fig:grid_of_images}, the agreement of azimuth estimates within the specimen areas emphasizes the effectiveness of our transformation framework. \par
To quantify whether our augmentation preserves the physical admissibility of the Mueller matrices, we randomly sampled 100 pixel coordinate pairs in each of 19 test images and computed the point correspondences after applying our polar-aware transform. We evaluated admissibility at each location using the characteristic polynomial~\cite{novikova2024time} both before and after rotation. From the 1654 coordinate pairs considered, 1642 remained admissible in both configurations and 12 became inadmissible after rotation; all other points lay outside the field of view and were excluded from the analysis. This yields an overall accuracy and sensitivity of 99.274~\%. Specificity cannot be calculated because every valid matrix was admissible prior to transformation. In contrast, a purely spatial transform without polar corrections preserved admissibility for all sampled points. These results demonstrate that our polar-aware augmentation maintains the physical integrity of the Mueller matrices with a negligible failure rate.
\subsection{Deep learning evaluation}
To validate our proposed augmentations for the intended use, we apply the polarimetric transformation framework to brain tumor tissue classification in our related study~\cite{hahne24muller}.
\subsubsection{Dataset}
%
Our dataset consists of monochromatic polarimetric images with a resolution of 388×516 pixels, acquired using a custom-built imaging system~\cite{Rodriguez-Nunez:21, Lindberg2019, schucht2020}. It includes images of 20 brain tumor samples obtained during neurosurgy and 19 tumor-free samples collected from autopsies. For clarity, we refer to tumor-free samples as \textit{healthy} throughout this paper. Neuropathologists annotated the data with pixel-level GT labels for 4 distinct tissue classes: tumor, healthy, white matter, and gray matter. %
For evaluation, we employ a 3-fold cross-validation protocol with a fixed validation set of 4 healthy and 5 tumor samples, held constant across all folds. The remaining 30 samples (15 healthy and 15 tumor) are evenly partitioned into 3 folds, each comprising 10 samples (5 per class). In each round, two folds are used for training (20 samples), and the remaining fold is used for testing (10 samples), such that each fold serves as the test set once.
\par
As the network input, we feed Mueller matrix data $\mathbf{M}_i$ and its augmented $\mathbf{M}'_i$ representation, respectively. 
Our approach follows the method described in~\cite{hahne24muller}. By doing so, the network can more effectively learn key features relevant to the segmentation task. 
\subsubsection{Training}
\begin{figure*}[ht]
    \centering
    \begin{minipage}[t]{.49\textwidth}
        \centering
        \includegraphics[width=\linewidth]{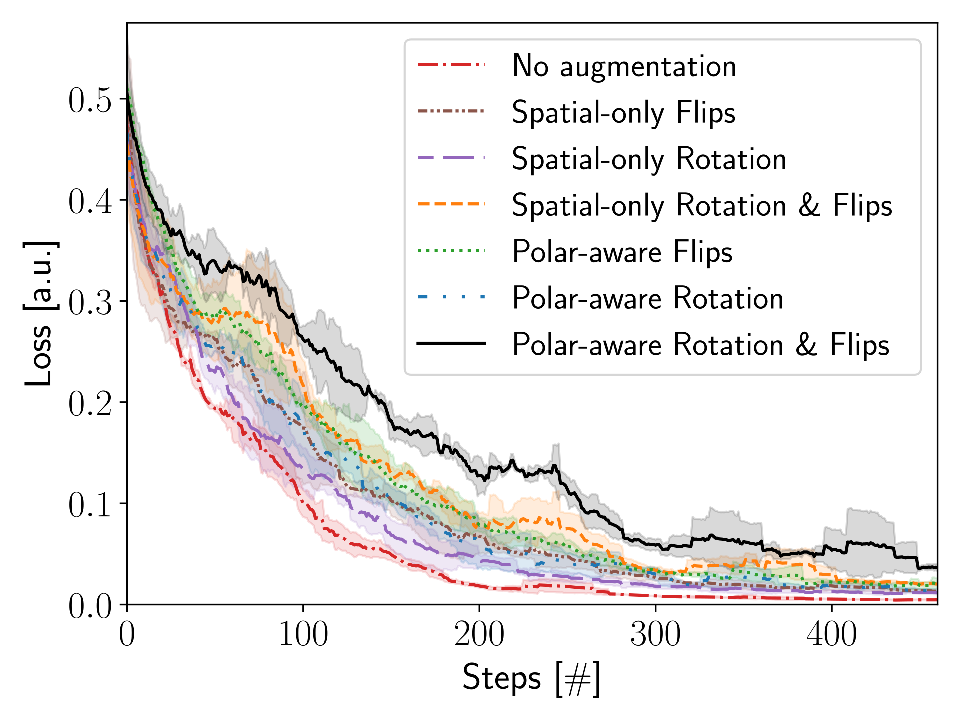}
        \subcaption[]{Training}
    \end{minipage}
    \hfill
    \begin{minipage}[t]{.49\textwidth}
        \centering 
        \includegraphics[width=\linewidth]{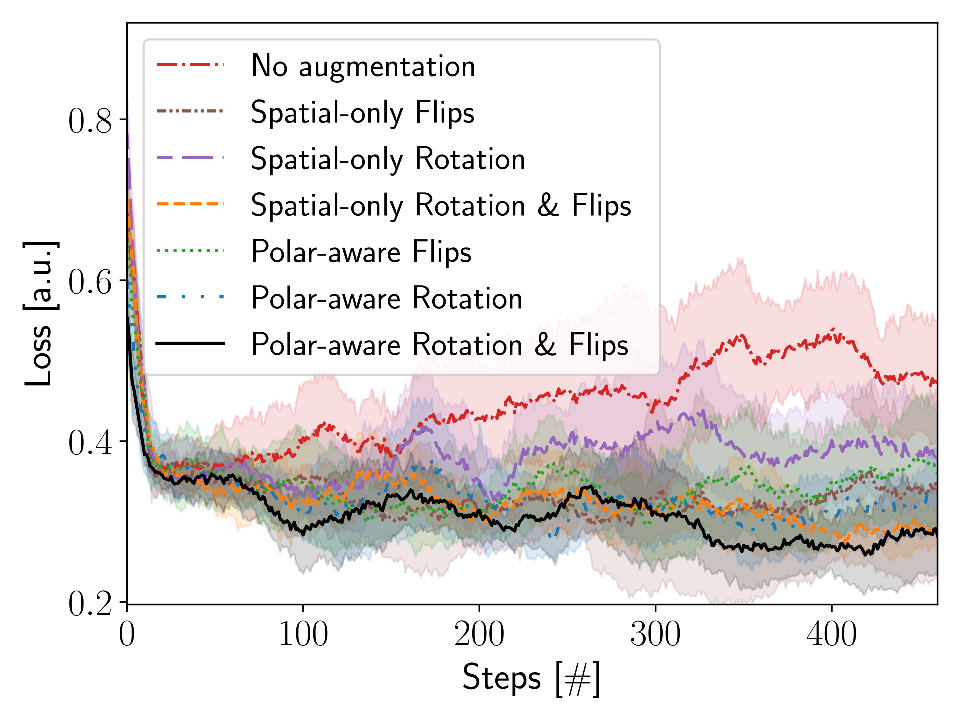}
        \subcaption[]{Validation\label{subfig:curves:b}}
    \end{minipage}
    \caption{\textbf{Training and validation curves} in the presence and absence of isometric transformations to augment Mueller matrix images. A stagnation or rise in the validation loss points to potential over-fitting to the training data. It can be seen in (\subref{subfig:curves:b}) that our proposed polarimetric rotation achieves the lowest average loss across all k-fold validation runs.}
    \label{fig:curves}
\end{figure*}

For training, we use the widely adopted U-Net, a fully convolutional network variant with multi-scale skip connections and learnable up-convolution layers, which has become a standard in medical image segmentation~\cite{unet2016ronneberger}. 
During training, each frame is randomly augmented. A rotation is applied with a 50\% chance, while horizontal and vertical flips are applied independently with 25\% probability each. The rotation angle, $\theta\in[-\pi/4,\pi/4)$, is selected from a uniform distribution over this interval. 
We train on an Nvidia RTX 4090 with a batch size 2 for 200 epochs using an initial learning rate of $1\mathrm{e}{-4}$, which converges to zero by cosine annealing. 
The training objective is a cross-entropy loss, commonly used in semantic segmentation~\cite{unet2016ronneberger}, to ensure accurate pixel-wise classification by minimizing the discrepancy between predicted and ground-truth class distributions.
Figure~\ref{fig:curves} depicts the learning progress by means of the loss curve plots. \par

Our training setup focuses on labeled pixels, specifically excluding labels from pixels that meet certain criteria to enhance the network's ability to distinguish between key tissue types. First, labels for background pixels are skipped, allowing the network to focus on differentiating between healthy and tumor regions in white and gray matter. Pixels with saturated intensities in the Mueller matrix channels remain unlabeled to prevent artifacts during training. %
The labels are reorganized into four distinct classes: healthy white matter, tumor white matter, healthy gray matter, and tumor gray matter. This allows to retain the option to revert to a simplified healthy vs. tumor or white vs. gray matter classification if needed. Given the primary medical interest in identifying tumors within white matter, the setup merges predicted logits and labels for healthy and tumor gray matter into a single class representing gray matter. To achieve this, healthy and tumor regions in gray matter are unified by setting \( y_{\text{gm}} = y_{\text{hgm}} \lor y_{\text{tgm}} \) for the labels and \( x_{\text{gm}} = \max(x_{\text{hgm}}, x_{\text{tgm}}) \) for predictions post-training, facilitating a practical focus on tumor identification in white matter regions.

%
In Fig.~\ref{fig:curves}, it can be seen that the validation loss fails to converge without augmentation, which indicates over-fitting at an early training stage. We overcome this problem by incorporating our proposed isometric transformations, which retain low validation loss values. Using all of our augmentations, we obtain the lowest overall validation loss toward the end of maximum training time. \par
%
In Table~\ref{tab:times}, we present CPU-based computation times for our proposed transformations based on 388 $\times$ 516 pixels per frame. There it is seen that applying transformations on raw calibration data introduces a computational overhead, as the number of feature channels is tripled. In fact, we measure an average frame processing time of 210~ms using our rotation of calibration data in Eqs.~\eqref{eq:mmext} and \eqref{eq:awext:rearrange} when embedded into the data loader. This adds roughly 200~ms compared to a polarimetric rotation on pre-computed Mueller matrix data as in Eq.~\eqref{eq:mmpolar}. However, data augmentation typically occurs concurrently with training, utilizing multiple CPUs, and is generally not regarded as time-critical. We provide spatial-only computation times to demonstrate the relative performance impact of additional polarimetric transformations.
\begin{table*}[htbp]
\centering
\caption{Avg. computation times $\pm$ standard deviation for our isometric transformation implementations.}
\begin{tabular}{r c c c c c c}
\toprule
\multicolumn{1}{l}{} & \multicolumn{3}{c}{\textbf{Rotation}} & \multicolumn{3}{c}{\textbf{Flip}} \\
& Eq.~\eqref{eq:mmpolar} & Eqs.~\eqref{eq:mmext} \& \eqref{eq:awext} & Eqs.~\eqref{eq:mmext} \& \eqref{eq:awext:rearrange} & Eq.~\eqref{eq:mmpolar} & Eqs.~\eqref{eq:mmext} \& \eqref{eq:awext} & Eqs.~\eqref{eq:mmext} \& \eqref{eq:awext:rearrange} \\
\cmidrule(lr){2-4}\cmidrule(lr){5-7}
\textbf{Spatial-only} & 15 $\pm$ 3~ms & 46 $\pm$ 5~ms & 46 $\pm$ 5~ms & 1 $\pm$ 0~ms & 8 $\pm$ 1~ms & 8 $\pm$ 1~ms \\
\textbf{Proposed} & 24 $\pm$ 4~ms & 347 $\pm$ 62~ms & 210 $\pm$ 37~ms & 12 $\pm$ 2~ms & 121 $\pm$ 24~ms & 39 $\pm$ 5~ms \\
\bottomrule\bottomrule
\end{tabular}
\label{tab:times}
\end{table*}

\subsubsection{Test}
To explore our algorithms' impact on unseen data, we provide a quantitative assessment on the image test set. While we augment images during training, all test images remain untreated. 
For classification performance, we report the Dice similarity coefficient~(DSC) and Intersection over Union~(IoU) between tumor and healthy white matter as well as gray matter tissue types. 
For test evaluation, we select the best model based on the highest validation DSC across all epochs. %
%
Table~\ref{tab:segment:scores} shows these semantic segmentation metrics in the presence and absence of our proposed transformations, compared against naive augmentation methods for baseline context. This includes spatial-only rotations and flips versions unaware of Eqs.~\ref{eq:mmext}. To further demonstrate the impact of our method relative to standard approaches, we apply a 50\% probability for adding element-wise noise from a normal distribution $\mathcal{N}(0, 0.1)$ to each $\mathbf{M}'_i$ in the training data. 
Finally, the classification results without any augmentations are provided as an ablation study to highlight the contributions of each augmentation type. \par

\begin{table}[h!]
    \centering
    \caption{Semantic image segmentation scores (IoU, DSC), showing improved results with polar-aware augmentations.}
    \begin{tabular}{llccc}
    \toprule
    Augment. Type & Method & IoU $\pm$ std $\uparrow$ & DSC $\pm$ std $\uparrow$ \\
    \midrule
    No augmentation & Baseline & 0.784 $\pm$ 0.060 & 0.863 $\pm$ 0.055 \\
    Gaussian noise & $\mathcal{N}(0,0.1)$ & 0.719 $\pm$ 0.050 & 0.814 $\pm$ 0.046 \\
    \midrule
    \multirow{2}{*}{Flipping} & Spatial-only & 0.775 $\pm$ 0.057 & 0.855 $\pm$ 0.055 \\
                              & Polar-aware & 0.799 $\pm$ 0.035 & 0.872 $\pm$ 0.039 \\
    \midrule
    \multirow{2}{*}{Rotations} & Spatial-only & 0.786 $\pm$ 0.052 & 0.866 $\pm$ 0.046 \\
                                & Polar-aware & 0.800 $\pm$ 0.055 & 0.873 $\pm$ 0.049 \\
    \midrule
    \multirow{2}{*}{Flip \& Rotations} & Spatial-only & 0.780 $\pm$ 0.061 & 0.860 $\pm$ 0.057 \\
                                & Polar-aware & 0.813 $\pm$ 0.034 & 0.882 $\pm$ 0.037 \\
    \bottomrule
    \bottomrule
    \end{tabular}
    \label{tab:segment:scores}
\end{table}

\begin{figure*}[h!]
\centering
\begin{minipage}[t]{\textwidth}
    \centering
    \includegraphics[width=.8\textwidth]{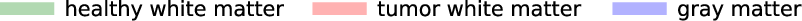}
\end{minipage}
\\[1em]
\begin{minipage}[t]{0.1\textwidth}
\raisebox{2.55\height}{\parbox{\textwidth}{\raggedright Absence}}
\end{minipage}
\hfill
\begin{minipage}[b]{0.0825\textwidth}
\centering
\includegraphics[width=\textwidth]{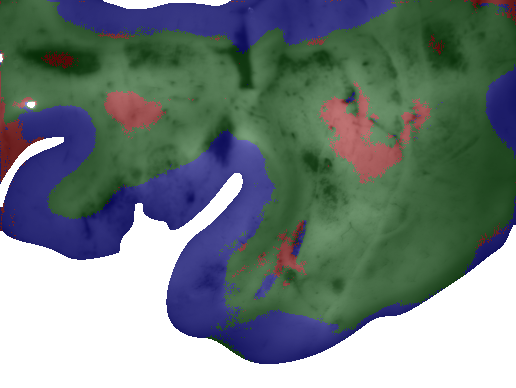}\\
\end{minipage}
\begin{minipage}[b]{0.0825\textwidth}
\centering
\includegraphics[width=\textwidth]{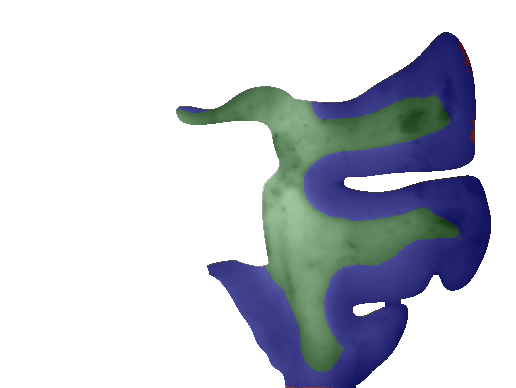}\\
\end{minipage}
\begin{minipage}[b]{0.0825\textwidth}
\centering
\includegraphics[width=\textwidth]{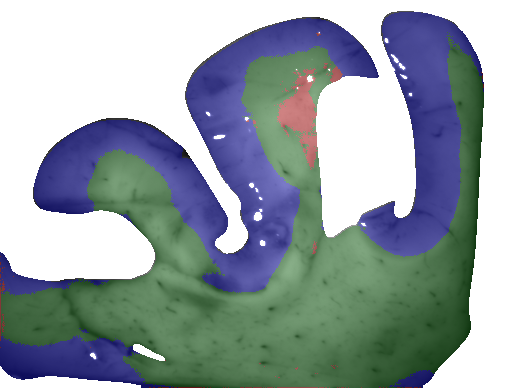}\\
\end{minipage}
\begin{minipage}[b]{0.0825\textwidth}
\centering
\includegraphics[width=\textwidth]{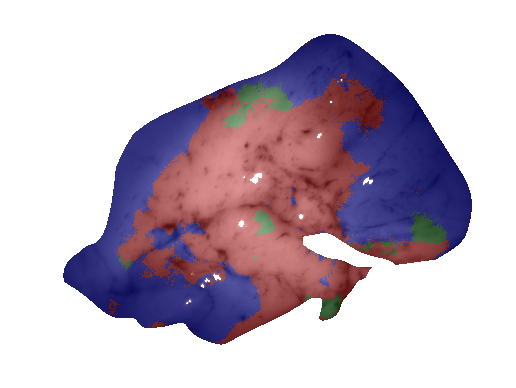}\\
\end{minipage}
\begin{minipage}[b]{0.0825\textwidth}
\centering
\includegraphics[width=\textwidth]{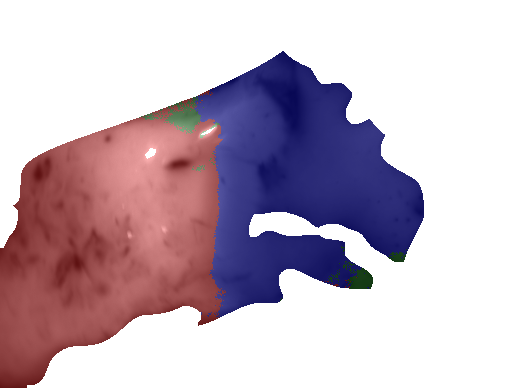}\\
\end{minipage}
\begin{minipage}[b]{0.0825\textwidth}
\centering
\includegraphics[width=\textwidth]{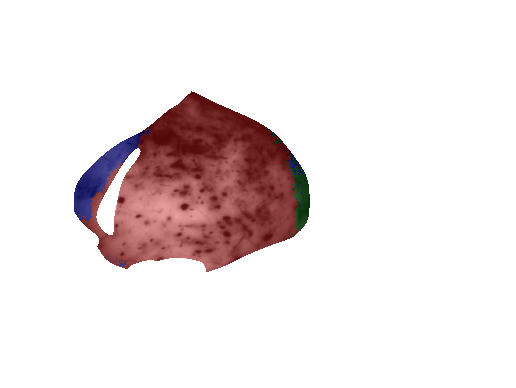}\\
\end{minipage}
\begin{minipage}[b]{0.0825\textwidth}
\centering
\includegraphics[width=\textwidth]{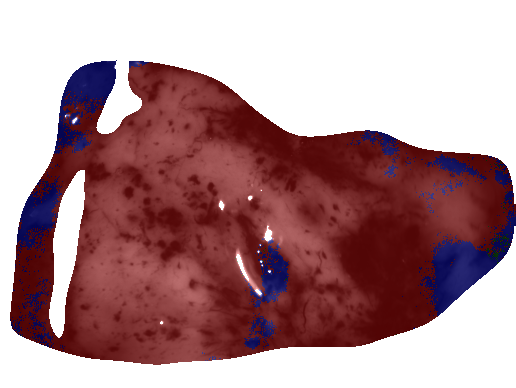}\\
\end{minipage}
\begin{minipage}[b]{0.0825\textwidth}
\centering
\includegraphics[width=\textwidth]{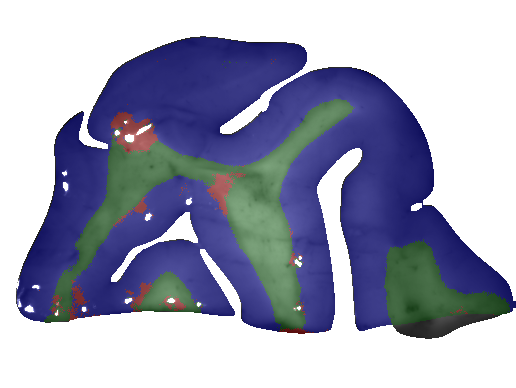}\\
\end{minipage}
\begin{minipage}[b]{0.0825\textwidth}
\centering
\includegraphics[width=\textwidth]{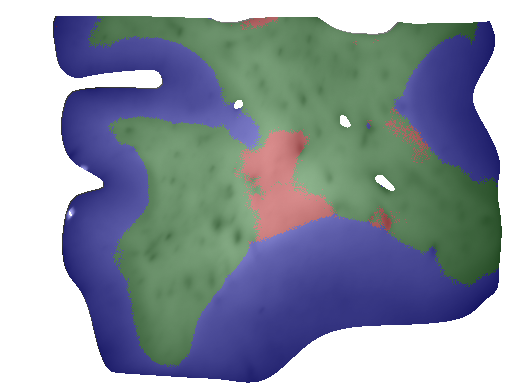}\\
\end{minipage}
\begin{minipage}[b]{0.0825\textwidth}
\centering
\includegraphics[width=\textwidth]{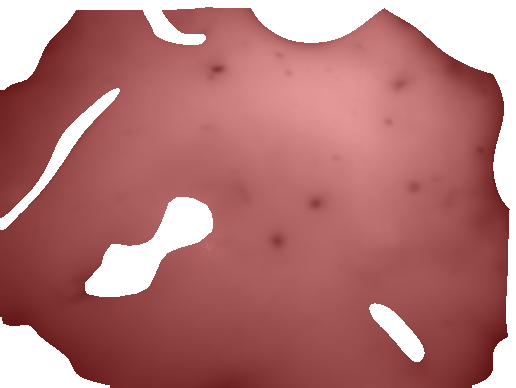}\\
\end{minipage}
\\[1em]
\begin{minipage}[t]{0.1\textwidth}
\raisebox{2.55\height}{\parbox{\textwidth}{\raggedright Noise}}
\end{minipage}
\hfill
\begin{minipage}[b]{0.0825\textwidth}
\centering
\includegraphics[width=\textwidth]{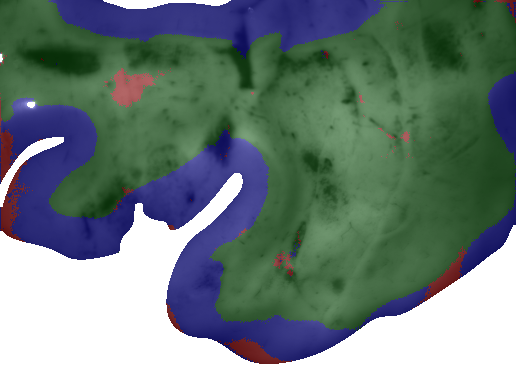}\\
\end{minipage}
\begin{minipage}[b]{0.0825\textwidth}
\centering
\includegraphics[width=\textwidth]{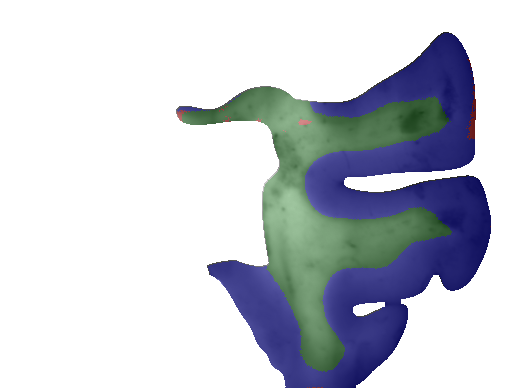}\\
\end{minipage}
\begin{minipage}[b]{0.0825\textwidth}
\centering
\includegraphics[width=\textwidth]{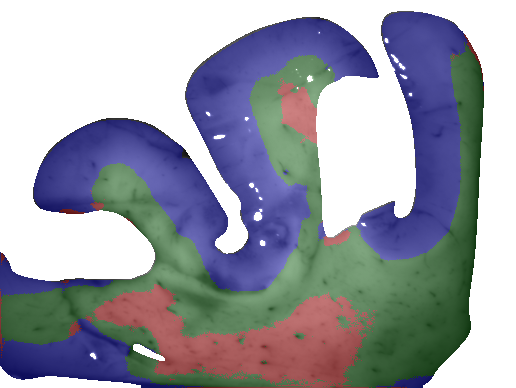}\\
\end{minipage}
\begin{minipage}[b]{0.0825\textwidth}
\centering
\includegraphics[width=\textwidth]{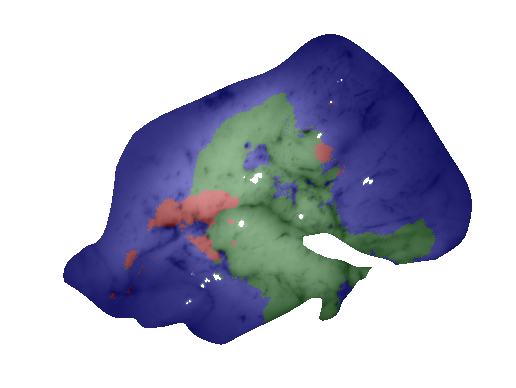}\\
\end{minipage}
\begin{minipage}[b]{0.0825\textwidth}
\centering
\includegraphics[width=\textwidth]{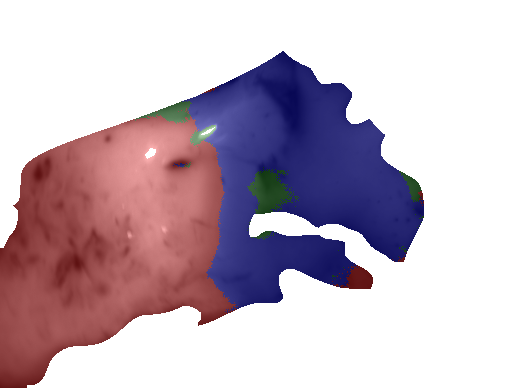}\\
\end{minipage}
\begin{minipage}[b]{0.0825\textwidth}
\centering
\includegraphics[width=\textwidth]{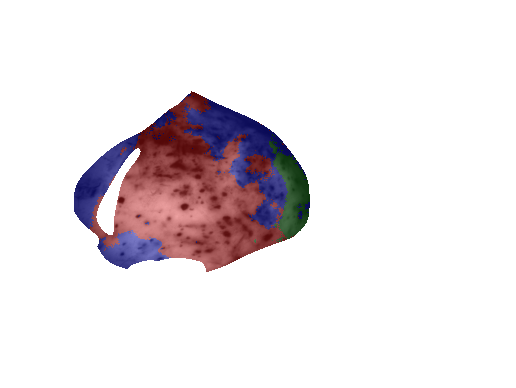}\\
\end{minipage}
\begin{minipage}[b]{0.0825\textwidth}
\centering
\includegraphics[width=\textwidth]{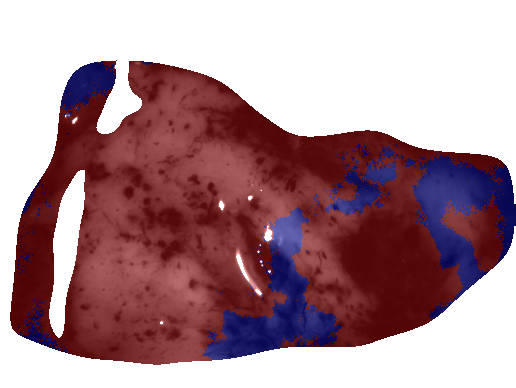}\\
\end{minipage}
\begin{minipage}[b]{0.0825\textwidth}
\centering
\includegraphics[width=\textwidth]{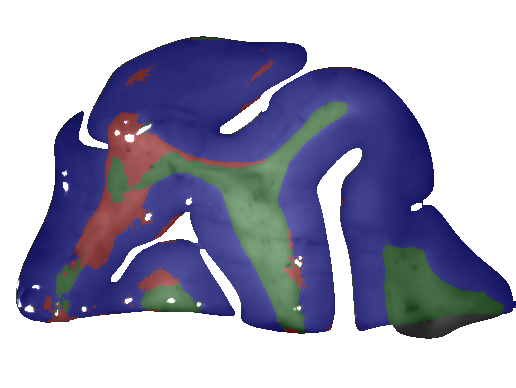}\\
\end{minipage}
\begin{minipage}[b]{0.0825\textwidth}
\centering
\includegraphics[width=\textwidth]{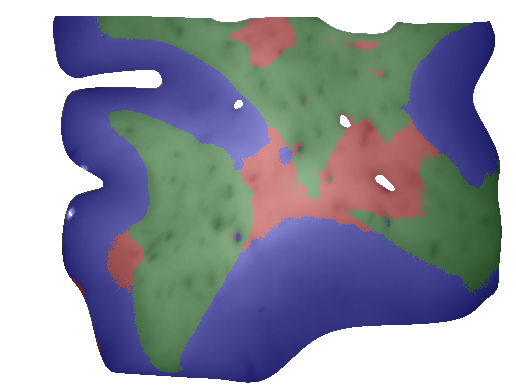}\\
\end{minipage}
\begin{minipage}[b]{0.0825\textwidth}
\centering
\includegraphics[width=\textwidth]{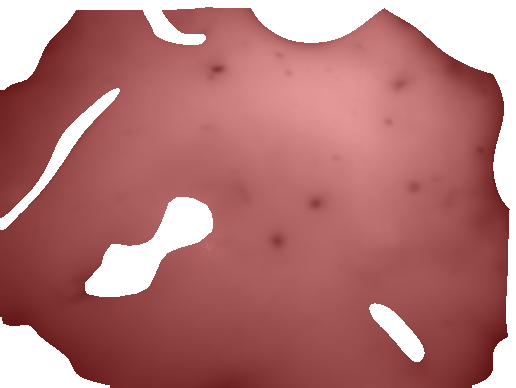}\\
\end{minipage}
\\[1em]
\begin{minipage}[t]{0.1\textwidth}
\raisebox{1.05\height}{\parbox{\textwidth}{\raggedright Spatial-only Flipping}}
\end{minipage}
\hfill
\begin{minipage}[b]{0.0825\textwidth}
\centering
\includegraphics[width=\textwidth]{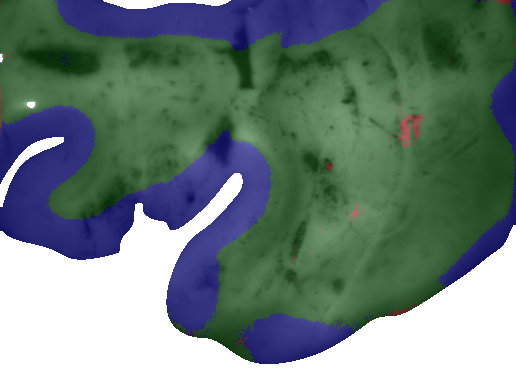}\\
\end{minipage}
\begin{minipage}[b]{0.0825\textwidth}
\centering
\includegraphics[width=\textwidth]{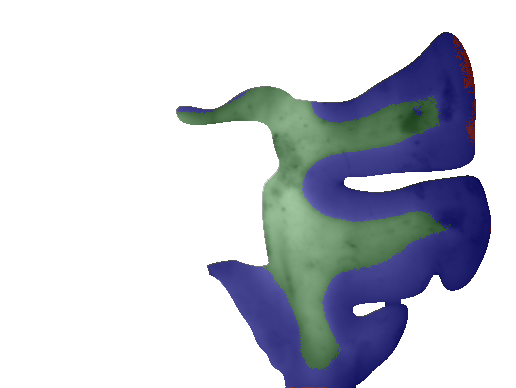}\\
\end{minipage}
\begin{minipage}[b]{0.0825\textwidth}
\centering
\includegraphics[width=\textwidth]{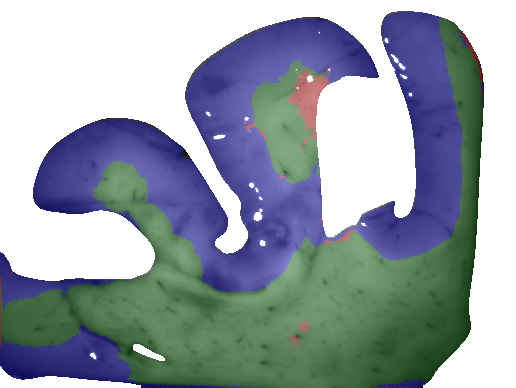}\\
\end{minipage}
\begin{minipage}[b]{0.0825\textwidth}
\centering
\includegraphics[width=\textwidth]{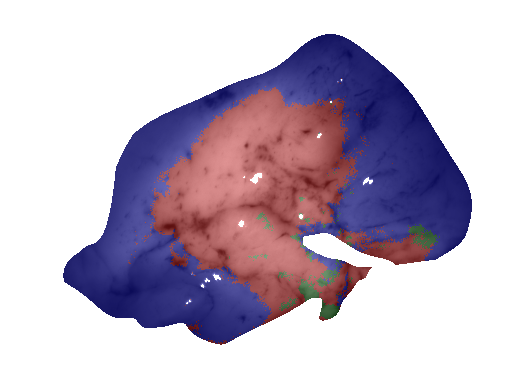}\\
\end{minipage}
\begin{minipage}[b]{0.0825\textwidth}
\centering
\includegraphics[width=\textwidth]{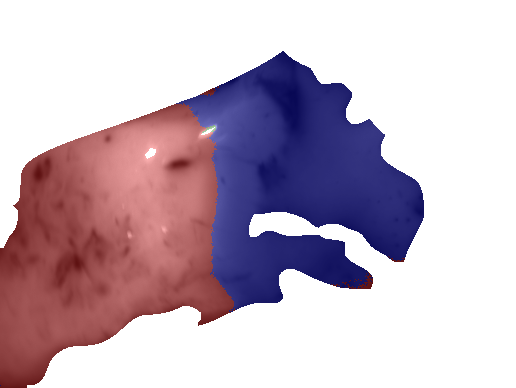}\\
\end{minipage}
\begin{minipage}[b]{0.0825\textwidth}
\centering
\includegraphics[width=\textwidth]{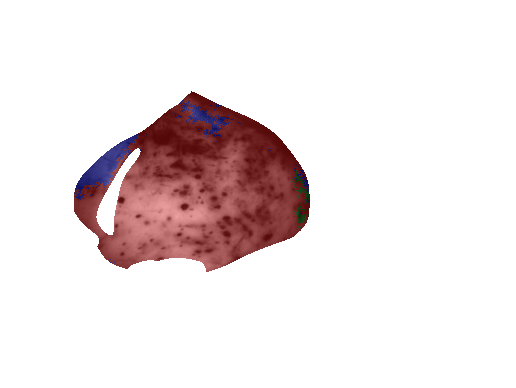}\\
\end{minipage}
\begin{minipage}[b]{0.0825\textwidth}
\centering
\includegraphics[width=\textwidth]{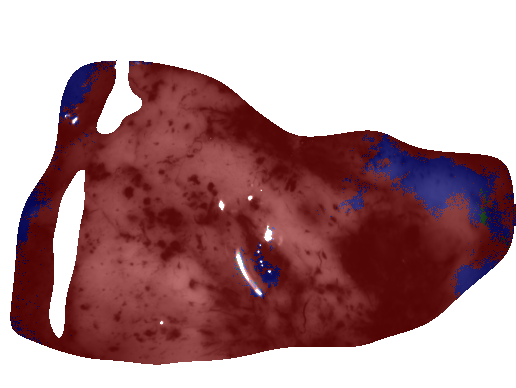}\\
\end{minipage}
\begin{minipage}[b]{0.0825\textwidth}
\centering
\includegraphics[width=\textwidth]{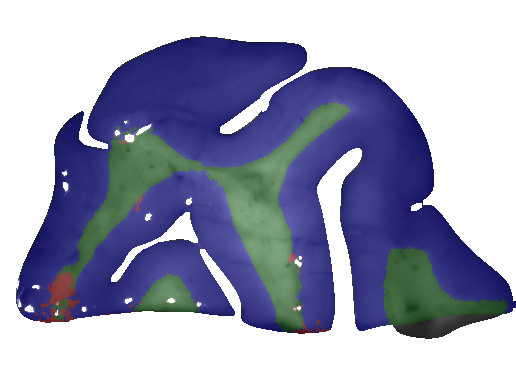}\\
\end{minipage}
\begin{minipage}[b]{0.0825\textwidth}
\centering
\includegraphics[width=\textwidth]{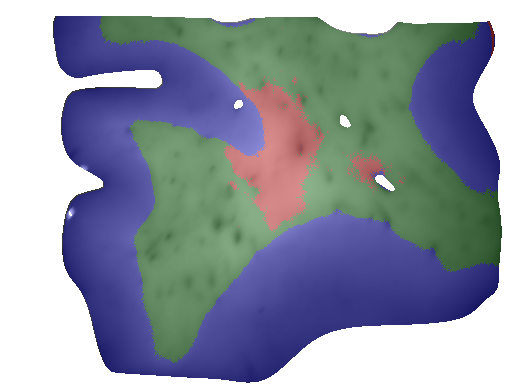}\\
\end{minipage}
\begin{minipage}[b]{0.0825\textwidth}
\centering
\includegraphics[width=\textwidth]{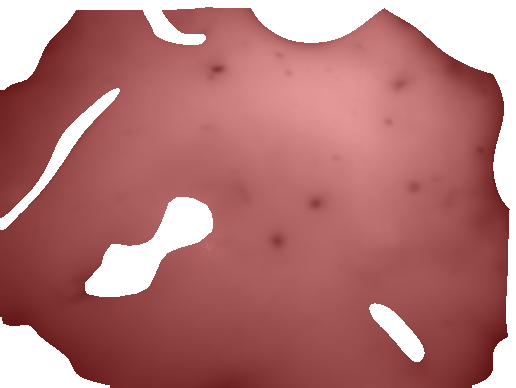}\\
\end{minipage}
\\[1em]
\begin{minipage}[t]{0.1\textwidth}
\raisebox{1.05\height}{\parbox{\textwidth}{\raggedright Spatial-only Rotation}}
\end{minipage}
\hfill
\begin{minipage}[b]{0.0825\textwidth}
\centering
\includegraphics[width=\textwidth]{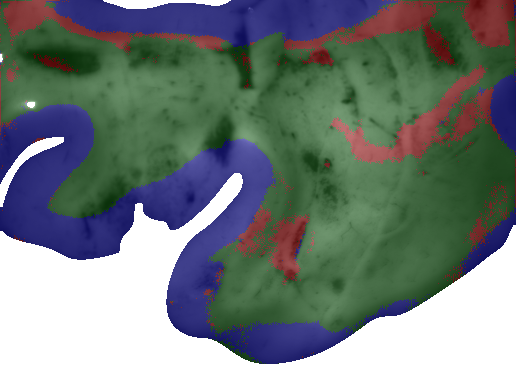}\\
\end{minipage}
\begin{minipage}[b]{0.0825\textwidth}
\centering
\includegraphics[width=\textwidth]{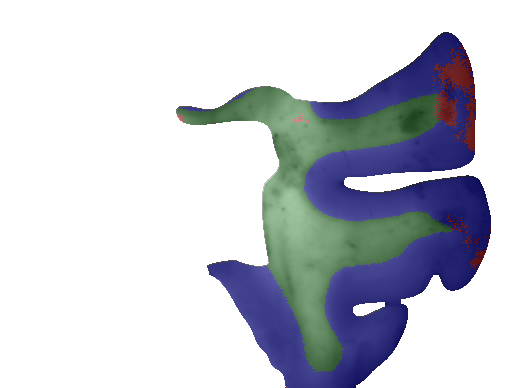}\\
\end{minipage}
\begin{minipage}[b]{0.0825\textwidth}
\centering
\includegraphics[width=\textwidth]{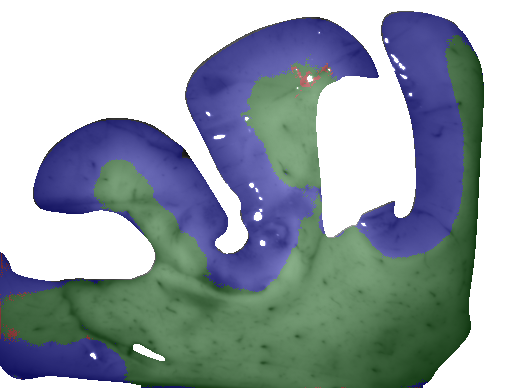}\\
\end{minipage}
\begin{minipage}[b]{0.0825\textwidth}
\centering
\includegraphics[width=\textwidth]{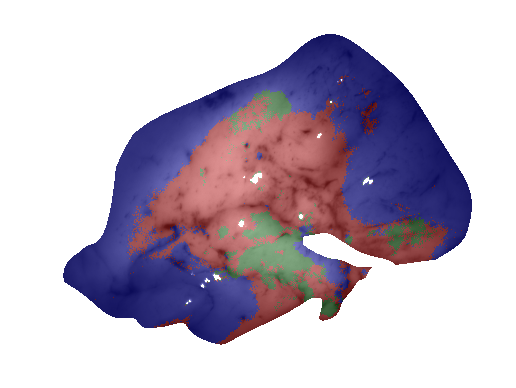}\\
\end{minipage}
\begin{minipage}[b]{0.0825\textwidth}
\centering
\includegraphics[width=\textwidth]{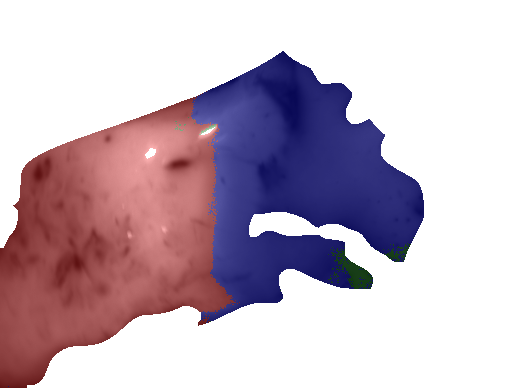}\\
\end{minipage}
\begin{minipage}[b]{0.0825\textwidth}
\centering
\includegraphics[width=\textwidth]{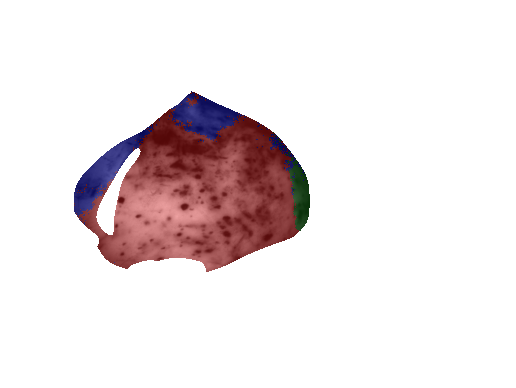}\\
\end{minipage}
\begin{minipage}[b]{0.0825\textwidth}
\centering
\includegraphics[width=\textwidth]{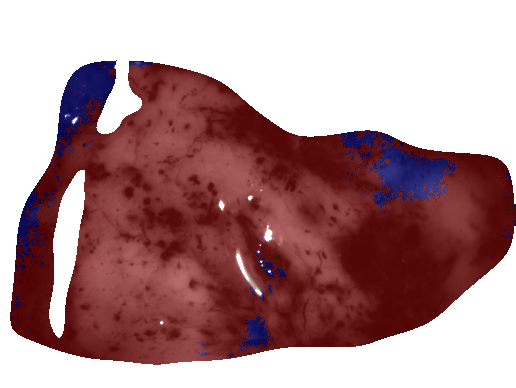}\\
\end{minipage}
\begin{minipage}[b]{0.0825\textwidth}
\centering
\includegraphics[width=\textwidth]{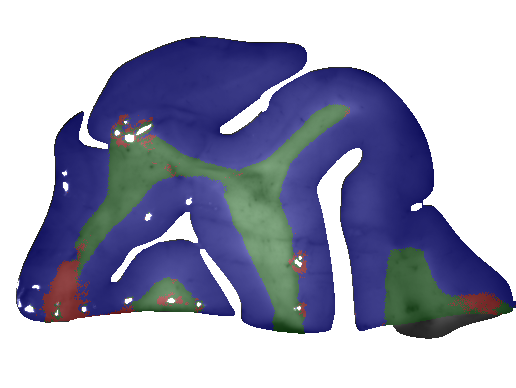}\\
\end{minipage}
\begin{minipage}[b]{0.0825\textwidth}
\centering
\includegraphics[width=\textwidth]{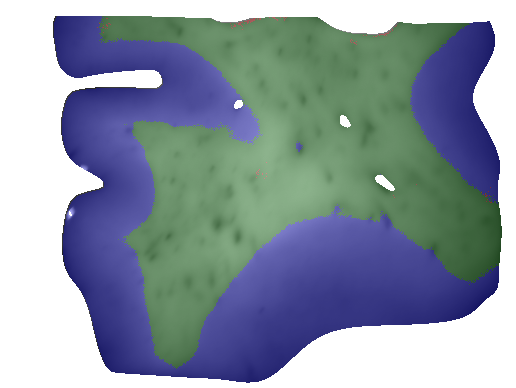}\\
\end{minipage}
\begin{minipage}[b]{0.0825\textwidth}
\centering
\includegraphics[width=\textwidth]{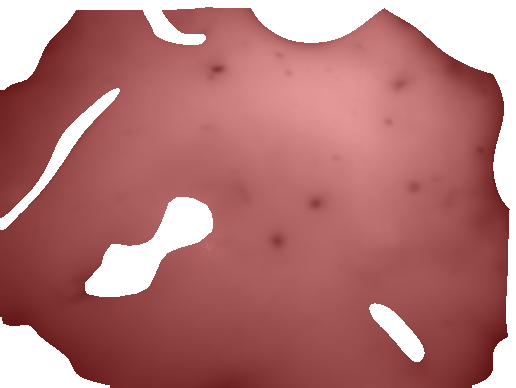}\\
\end{minipage}
\\[1em]
\begin{minipage}[t]{0.1\textwidth}
\raisebox{1.05\height}{\parbox{\textwidth}{\raggedright Spatial-only Rota \& Flip}}
\end{minipage}
\hfill
\begin{minipage}[b]{0.0825\textwidth}
\centering
\includegraphics[width=\textwidth]{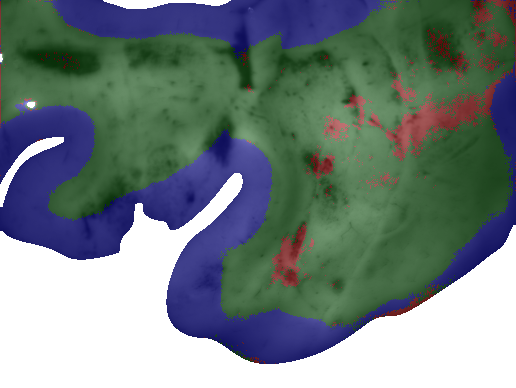}\\
\end{minipage}
\begin{minipage}[b]{0.0825\textwidth}
\centering
\includegraphics[width=\textwidth]{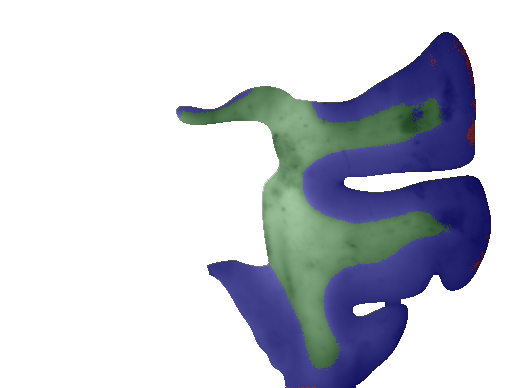}\\
\end{minipage}
\begin{minipage}[b]{0.0825\textwidth}
\centering
\includegraphics[width=\textwidth]{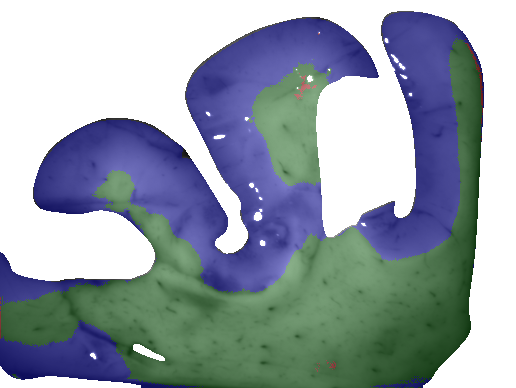}\\
\end{minipage}
\begin{minipage}[b]{0.0825\textwidth}
\centering
\includegraphics[width=\textwidth]{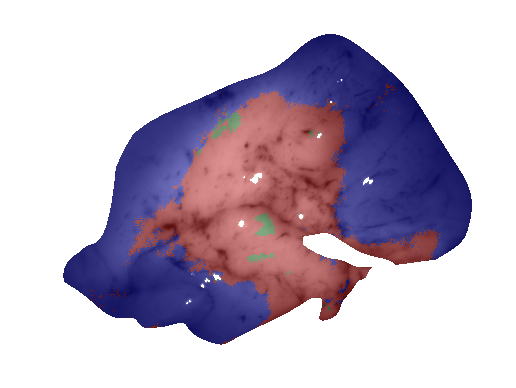}\\
\end{minipage}
\begin{minipage}[b]{0.0825\textwidth}
\centering
\includegraphics[width=\textwidth]{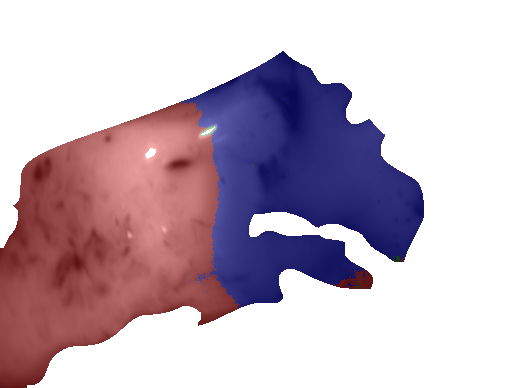}\\
\end{minipage}
\begin{minipage}[b]{0.0825\textwidth}
\centering
\includegraphics[width=\textwidth]{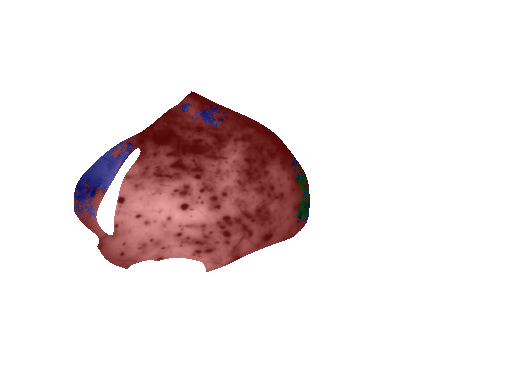}\\
\end{minipage}
\begin{minipage}[b]{0.0825\textwidth}
\centering
\includegraphics[width=\textwidth]{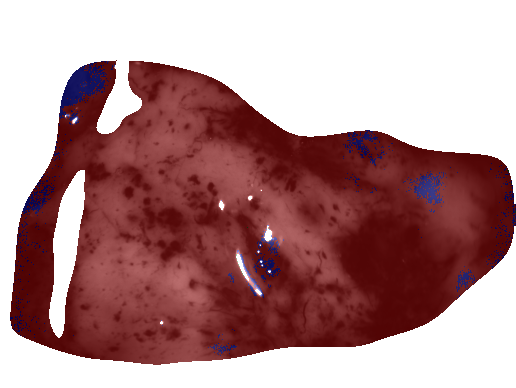}\\
\end{minipage}
\begin{minipage}[b]{0.0825\textwidth}
\centering
\includegraphics[width=\textwidth]{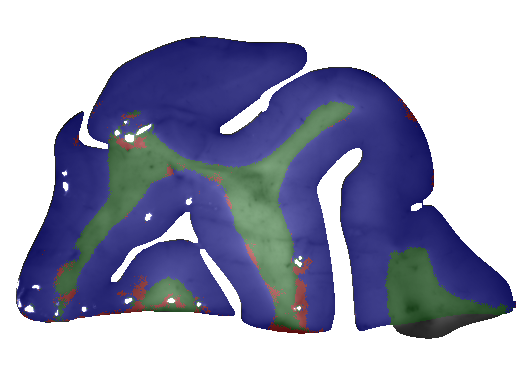}\\
\end{minipage}
\begin{minipage}[b]{0.0825\textwidth}
\centering
\includegraphics[width=\textwidth]{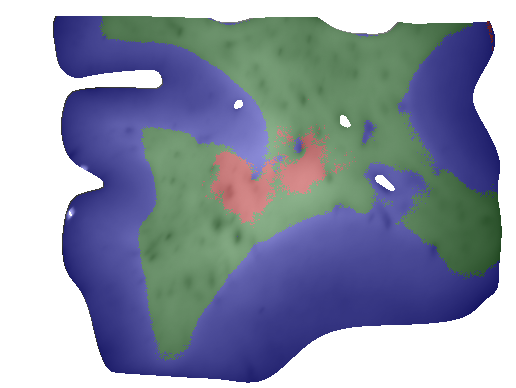}\\
\end{minipage}
\begin{minipage}[b]{0.0825\textwidth}
\centering
\includegraphics[width=\textwidth]{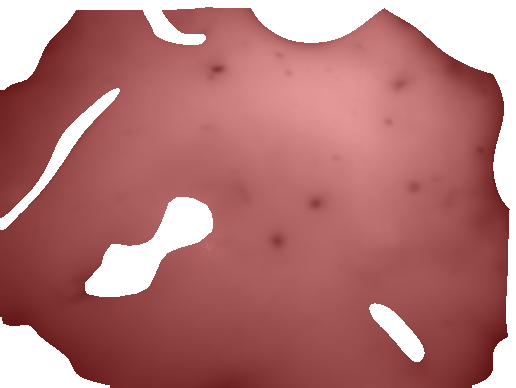}\\
\end{minipage}
\\[1em]
\begin{minipage}[t]{0.1\textwidth}
\raisebox{1.05\height}{\parbox{\textwidth}{\raggedright Polar-aware Flipping}}
\end{minipage}
\hfill
\begin{minipage}[b]{0.0825\textwidth}
\centering
\includegraphics[width=\textwidth]{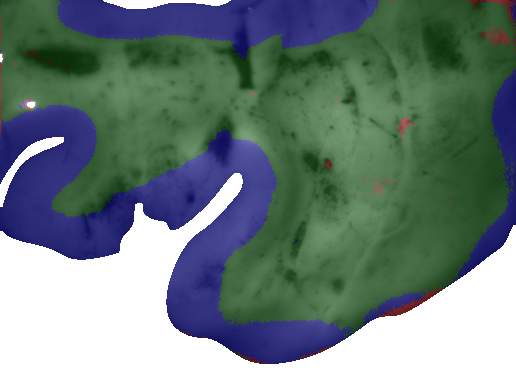}\\
\end{minipage}
\begin{minipage}[b]{0.0825\textwidth}
\centering
\includegraphics[width=\textwidth]{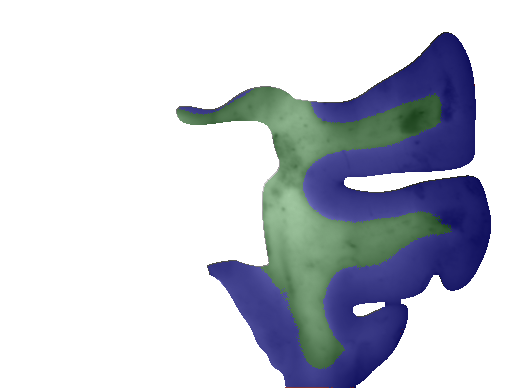}\\
\end{minipage}
\begin{minipage}[b]{0.0825\textwidth}
\centering
\includegraphics[width=\textwidth]{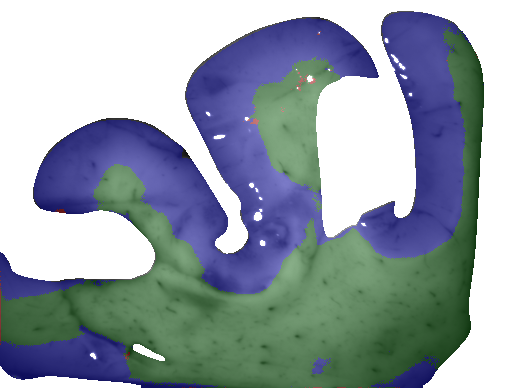}\\
\end{minipage}
\begin{minipage}[b]{0.0825\textwidth}
\centering
\includegraphics[width=\textwidth]{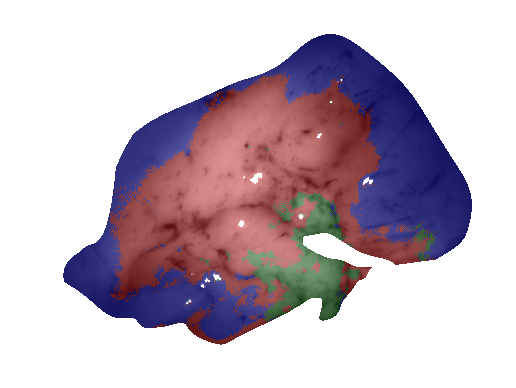}\\
\end{minipage}
\begin{minipage}[b]{0.0825\textwidth}
\centering
\includegraphics[width=\textwidth]{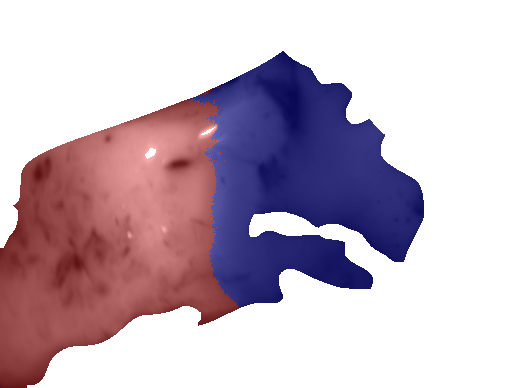}\\
\end{minipage}
\begin{minipage}[b]{0.0825\textwidth}
\centering
\includegraphics[width=\textwidth]{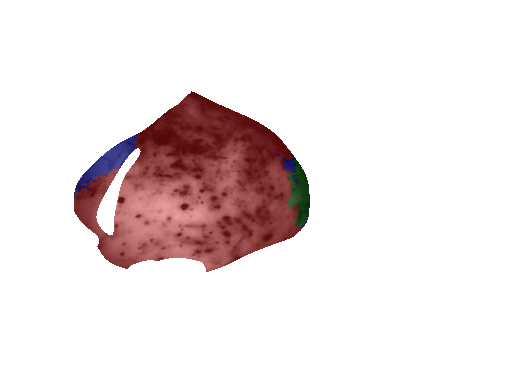}\\
\end{minipage}
\begin{minipage}[b]{0.0825\textwidth}
\centering
\includegraphics[width=\textwidth]{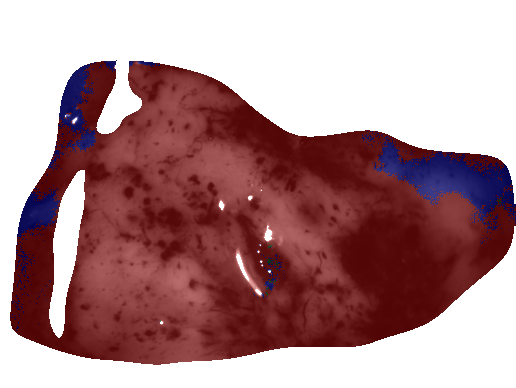}\\
\end{minipage}
\begin{minipage}[b]{0.0825\textwidth}
\centering
\includegraphics[width=\textwidth]{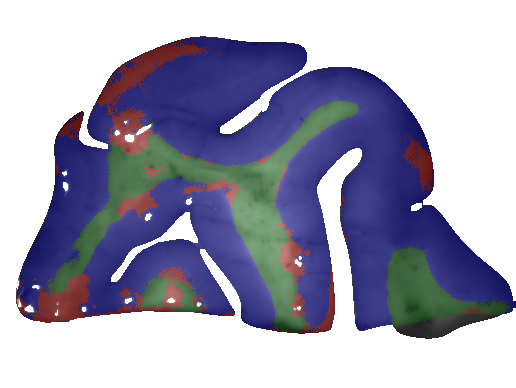}\\
\end{minipage}
\begin{minipage}[b]{0.0825\textwidth}
\centering
\includegraphics[width=\textwidth]{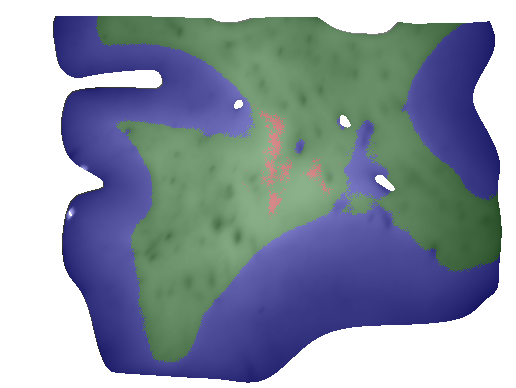}\\
\end{minipage}
\begin{minipage}[b]{0.0825\textwidth}
\centering
\includegraphics[width=\textwidth]{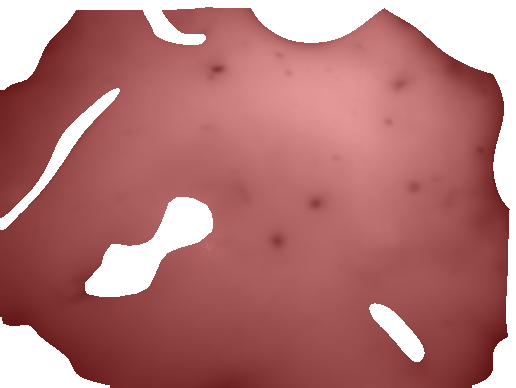}\\
\end{minipage}
\\[1em]
\begin{minipage}[t]{0.1\textwidth}
\raisebox{1.05\height}{\parbox{\textwidth}{\raggedright Polar-aware Rotation}}
\end{minipage}
\hfill
\begin{minipage}[b]{0.0825\textwidth}
\centering
\includegraphics[width=\textwidth]{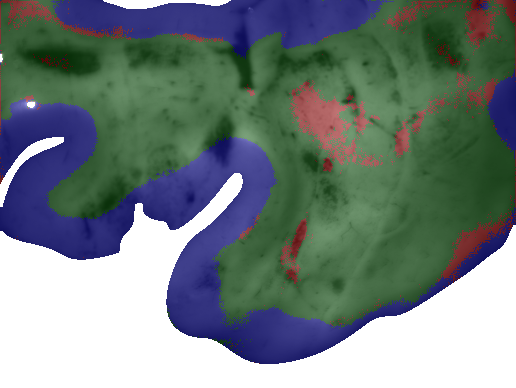}\\
\end{minipage}
\begin{minipage}[b]{0.0825\textwidth}
\centering
\includegraphics[width=\textwidth]{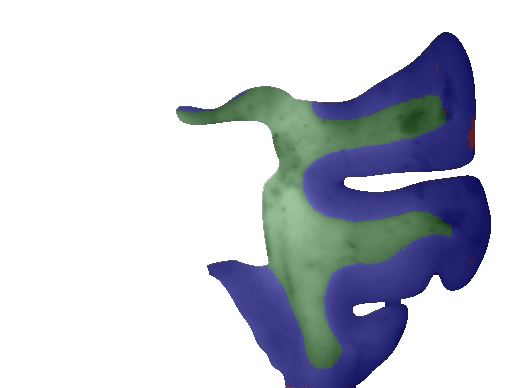}\\
\end{minipage}
\begin{minipage}[b]{0.0825\textwidth}
\centering
\includegraphics[width=\textwidth]{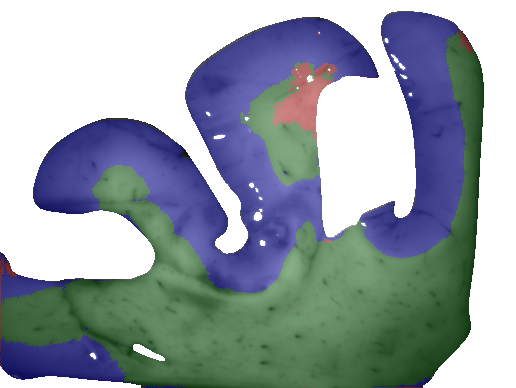}\\
\end{minipage}
\begin{minipage}[b]{0.0825\textwidth}
\centering
\includegraphics[width=\textwidth]{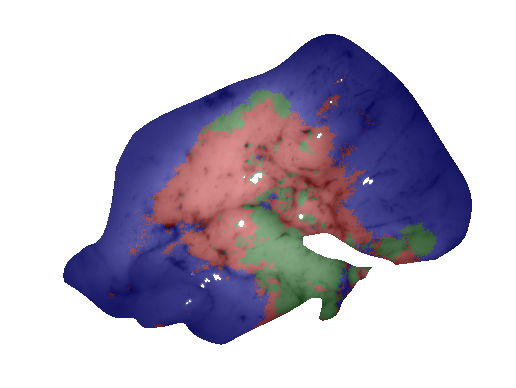}\\
\end{minipage}
\begin{minipage}[b]{0.0825\textwidth}
\centering
\includegraphics[width=\textwidth]{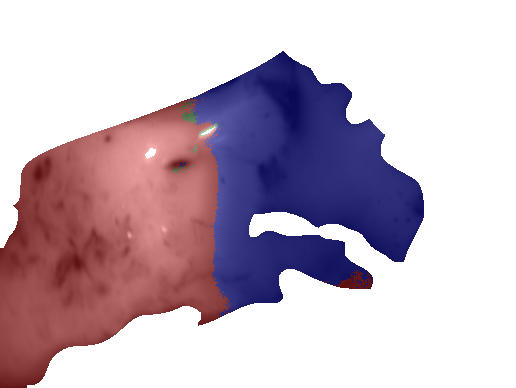}\\
\end{minipage}
\begin{minipage}[b]{0.0825\textwidth}
\centering
\includegraphics[width=\textwidth]{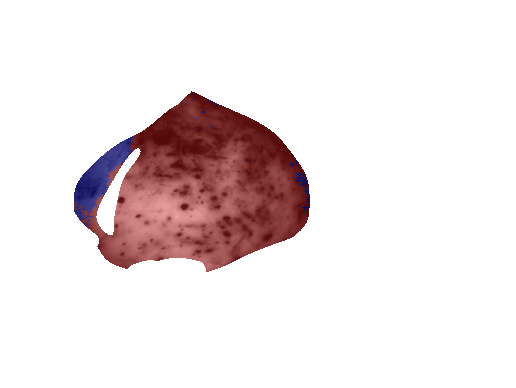}\\
\end{minipage}
\begin{minipage}[b]{0.0825\textwidth}
\centering
\includegraphics[width=\textwidth]{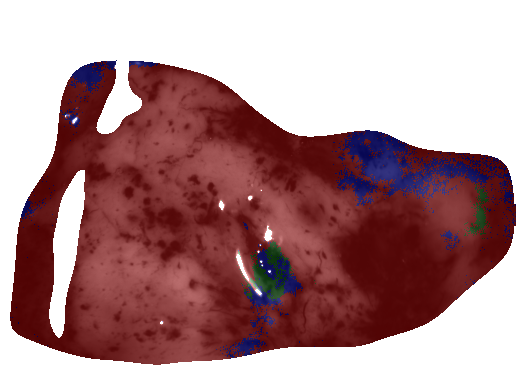}\\
\end{minipage}
\begin{minipage}[b]{0.0825\textwidth}
\centering
\includegraphics[width=\textwidth]{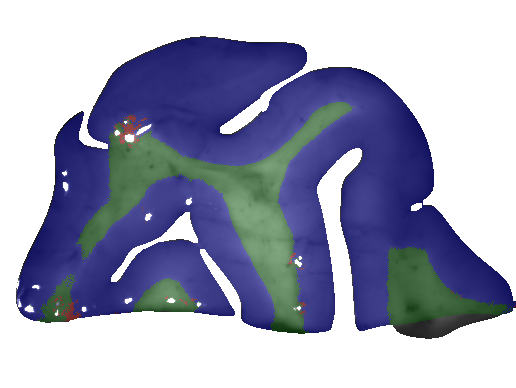}\\
\end{minipage}
\begin{minipage}[b]{0.0825\textwidth}
\centering
\includegraphics[width=\textwidth]{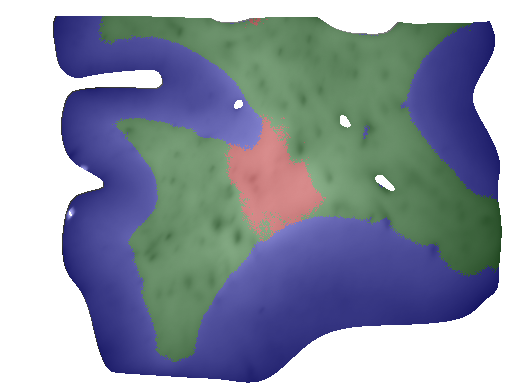}\\
\end{minipage}
\begin{minipage}[b]{0.0825\textwidth}
\centering
\includegraphics[width=\textwidth]{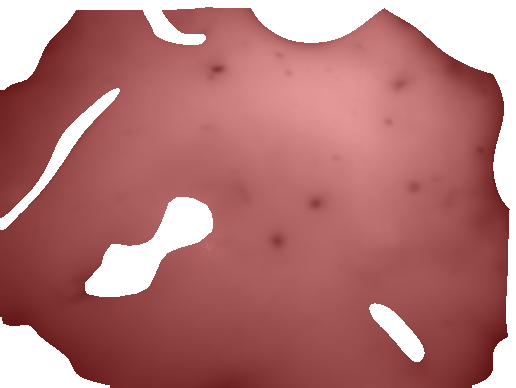}\\
\end{minipage}
\\[1em]
\begin{minipage}[t]{0.1\textwidth}
\raisebox{1.05\height}{\parbox{\textwidth}{\raggedright Polar-aware Rota \& Flip}}
\end{minipage}
\hfill
\begin{minipage}[b]{0.0825\textwidth}
\centering
\includegraphics[width=\textwidth]{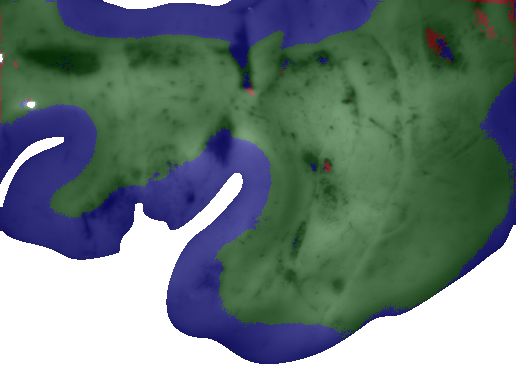}\\
\end{minipage}
\begin{minipage}[b]{0.0825\textwidth}
\centering
\includegraphics[width=\textwidth]{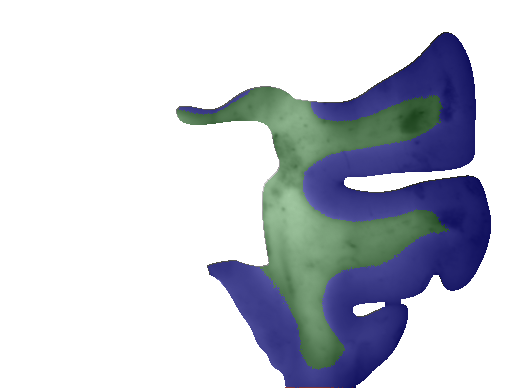}\\
\end{minipage}
\begin{minipage}[b]{0.0825\textwidth}
\centering
\includegraphics[width=\textwidth]{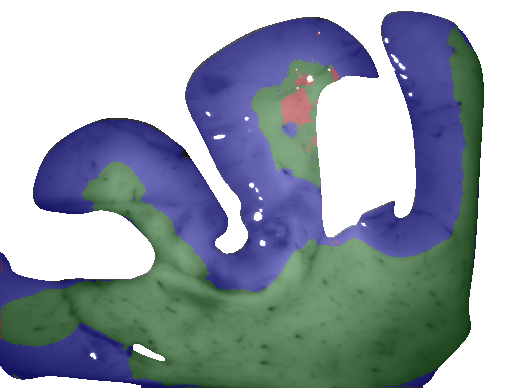}\\
\end{minipage}
\begin{minipage}[b]{0.0825\textwidth}
\centering
\includegraphics[width=\textwidth]{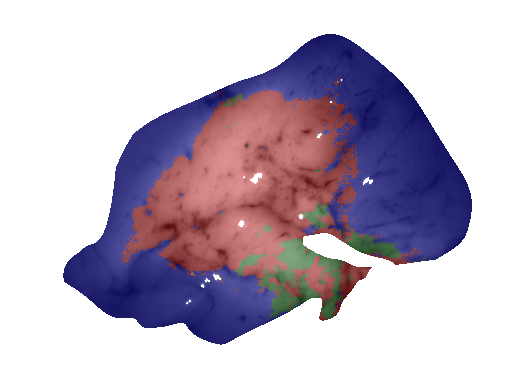}\\
\end{minipage}
\begin{minipage}[b]{0.0825\textwidth}
\centering
\includegraphics[width=\textwidth]{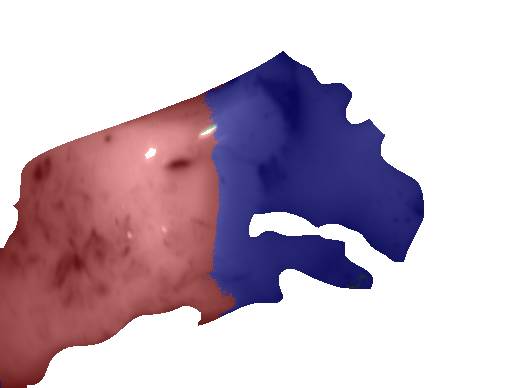}\\
\end{minipage}
\begin{minipage}[b]{0.0825\textwidth}
\centering
\includegraphics[width=\textwidth]{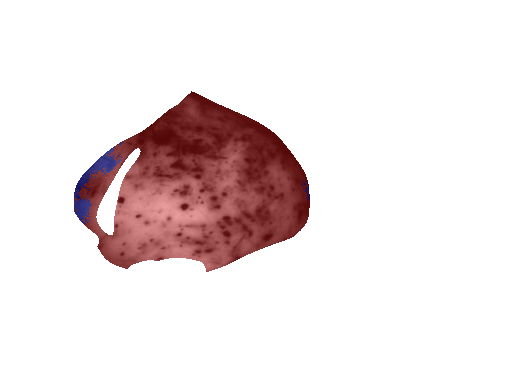}\\
\end{minipage}
\begin{minipage}[b]{0.0825\textwidth}
\centering
\includegraphics[width=\textwidth]{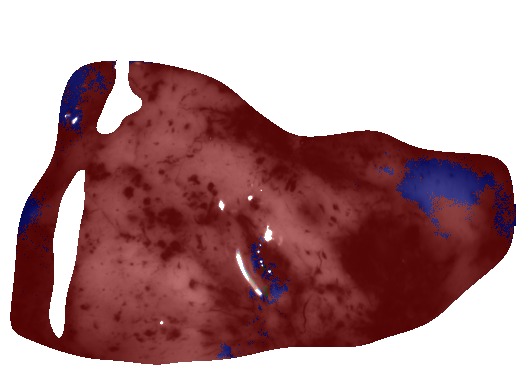}\\
\end{minipage}
\begin{minipage}[b]{0.0825\textwidth}
\centering
\includegraphics[width=\textwidth]{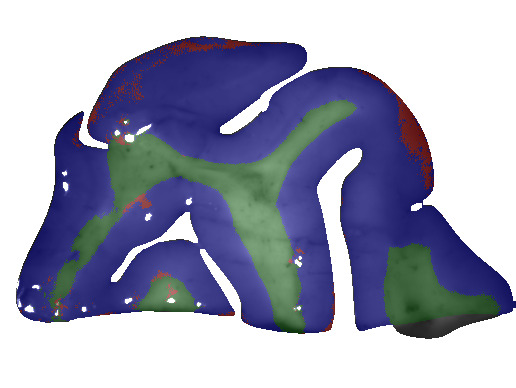}\\
\end{minipage}
\begin{minipage}[b]{0.0825\textwidth}
\centering
\includegraphics[width=\textwidth]{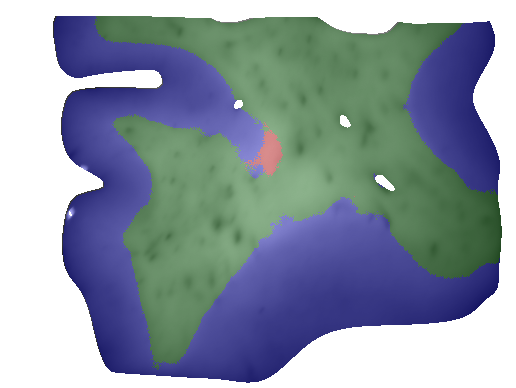}\\
\end{minipage}
\begin{minipage}[b]{0.0825\textwidth}
\centering
\includegraphics[width=\textwidth]{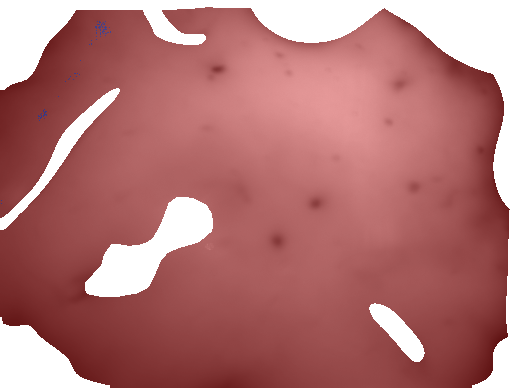}\\
\end{minipage}
\\[1em]
\begin{minipage}[t]{0.1\textwidth}
\raisebox{4.15\height}{\parbox{\textwidth}{\raggedright GT}}
\end{minipage}
\hfill
\begin{minipage}[b]{0.0825\textwidth}
\centering
\includegraphics[width=\textwidth]{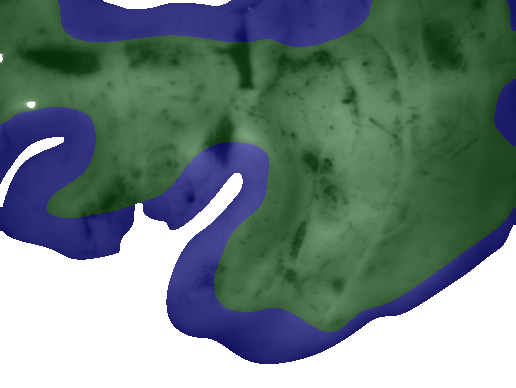}\\
healthy
\end{minipage}
\begin{minipage}[b]{0.0825\textwidth}
\centering
\includegraphics[width=\textwidth]{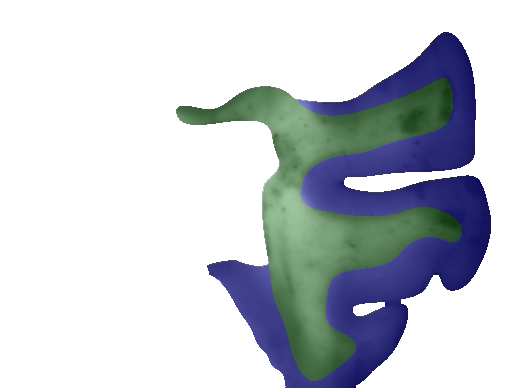}\\
healthy
\end{minipage}
\begin{minipage}[b]{0.0825\textwidth}
\centering
\includegraphics[width=\textwidth]{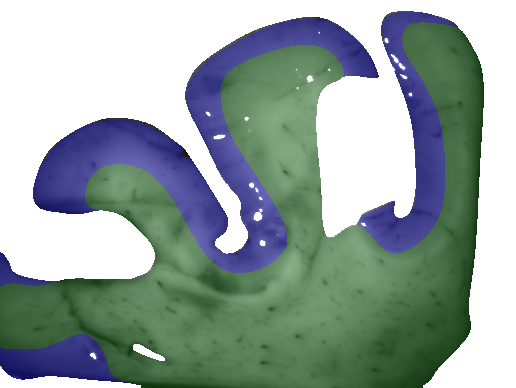}\\
healthy
\end{minipage}
\begin{minipage}[b]{0.0825\textwidth}
\centering
\includegraphics[width=\textwidth]{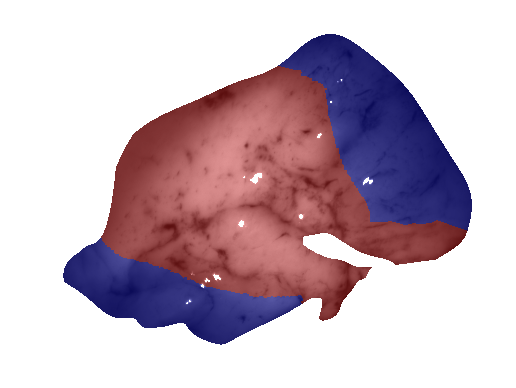}\\
O  3
\end{minipage}
\begin{minipage}[b]{0.0825\textwidth}
\centering
\includegraphics[width=\textwidth]{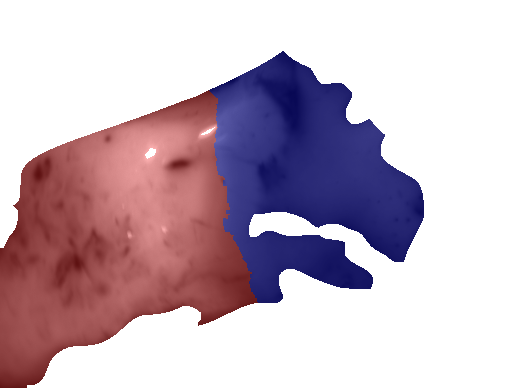}\\
A  4
\end{minipage}
\begin{minipage}[b]{0.0825\textwidth}
\centering
\includegraphics[width=\textwidth]{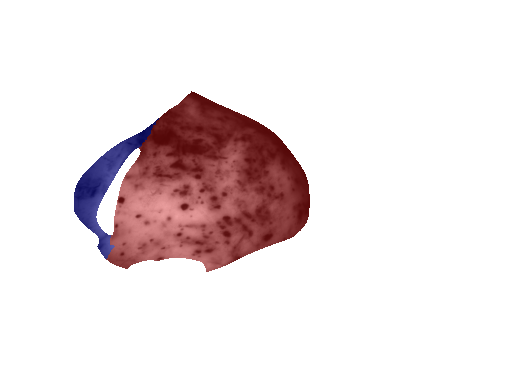}\\
GBM  4
\end{minipage}
\begin{minipage}[b]{0.0825\textwidth}
\centering
\includegraphics[width=\textwidth]{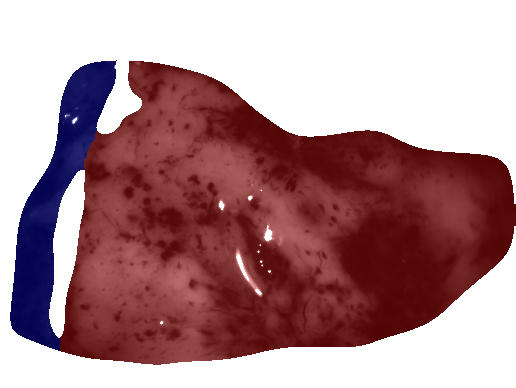}\\
GBM  4
\end{minipage}
\begin{minipage}[b]{0.0825\textwidth}
\centering
\includegraphics[width=\textwidth]{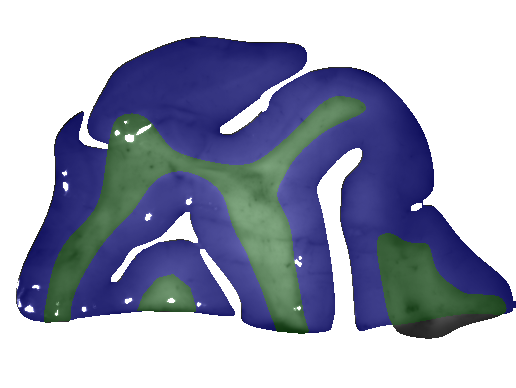}\\
healthy
\end{minipage}
\begin{minipage}[b]{0.0825\textwidth}
\centering
\includegraphics[width=\textwidth]{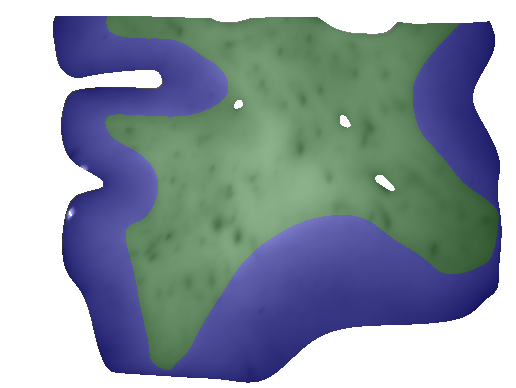}\\
healthy
\end{minipage}
\begin{minipage}[b]{0.0825\textwidth}
\centering
\includegraphics[width=\textwidth]{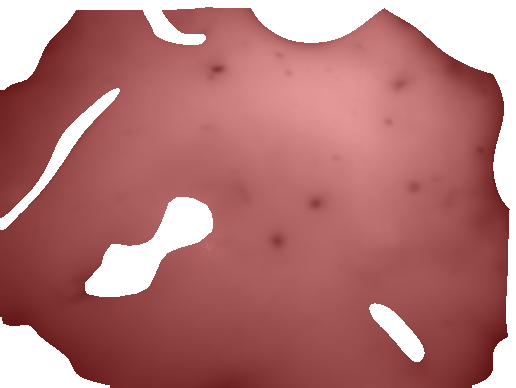}\\
A  2
\end{minipage}
\caption{\textbf{Semantic segmentation results from the test set.} The images show the intensity of $M_i^{(1,1)}$ from human brain samples overlaid with the predicted or annotated per-pixel class using an opacity of 30\%. The healthy examples come from autopsy and tumor cases were taken after surgery. Each image column corresponds to a different tissue sample while rows refer to the classification technique. The abbreviations for tumorous tissues are A = Astrocytoma, O = Oligodendroglioma, and GBM = Glioblastoma with numbers reflecting the grade according to the World Health Organization (WHO)~\cite{louis:2021}.}
\label{fig:segment}
\end{figure*}

The test scores in Table~\ref{tab:segment:scores} illustrate that our proposed polar-aware augmentation framework substantially enhances the segmentation performance on our brain tissue dataset. The absence of augmentations serves as baseline reference scores, against which all augmentations were evaluated. 
The Gaussian noise augmentation results in lower scores, potentially due to the disruptive effects on the polarimetric image details critical for segmentation. 
Conversely, spatial-only flipping and rotation show moderate gains over the baseline, but their polar-aware counterparts perform significantly better. For example, polar-aware flipping and rotation yield a significant larger DSC gain, outperforming spatial-only methods and indicating the advantage of preserving polarization properties in polarimetric data. This outcome aligns with the reduced overfitting observed in Fig.~\ref{fig:curves}, suggesting that our framework not only boosts segmentation accuracy but also enables better generalization across variations. These findings highlight the potential of physics-informed augmentations to improve performance, especially in settings with limited polarimetric data, thereby supporting the translational goal of polarimetric imaging for brain tissue differentiation. \par
%
For qualitative inspection, we provide segmented image results from the test set in Fig.~\ref{fig:segment}. The per-pixel class is determined by computing the $\operatorname{argmax}$ over the predicted class probabilities. 
We observe a general tendency of the baseline network to overpredict tumor regions, potentially as a consequence of the limited dataset size or class imbalance. However, we note that the bias is substantially mitigated when applying our proposed polar-aware augmentation strategies. The resulting segmentations exhibit improved spatial precision and reduced oversegmentation, consistent with the quantitative performance gains reported in Table~\ref{tab:segment:scores}. These outcomes underscore the value of embedding physical priors into the augmentation pipeline and demonstrate that even modest geometric transformations, when designed with polarization physics in mind, can yield meaningful improvements in model generalization. 
We now conclude with a broader reflection on the implications and limitations of our approach.
%

\section{Summary}
In this study, we developed a set of physics-based augmentation transforms specifically designed for Mueller matrix polarimetry, addressing the significant demands on restricted data availability and variety. Thereby, our study highlights the importance of special augmentation treatment when dealing with polarimetric images. Our findings demonstrate that traditional isometric transformations, such as standard rotations and flips, do not adequately account for the complexities of light polarization. By systematically modeling changes in the Mueller matrix, our approach ensures that augmentations are physically consistent with real-world conditions. \par %
This will have broader implications for the imaging community of Mueller matrix polarimetry. The special hardware, environmental setup and expert annotation make polarimetric data acquisition expensive such that the limitation in sample quantity will be a persisting challenge for scientists in the field. Our proposed framework offers a cost-effective solution to mimic data variety and facilitate more reliable polarimetric image learning in medical imaging and beyond. \par
The limitations of our study particularly lie in its focus on two-dimensional augmentation mappings. 
The proposed framework assumes a quasi-collinear configuration as misalignments between the illumination direction, detection axes, and the specimen surface can break the rotational invariance of our polarimetric transform. While this is analogous to specular highlights violating intensity invariance in conventional imaging, it can potentially further introduce angle-dependent polarization artifacts. \par %
Future research should explore more advanced simulation-based augmentations, including \mbox{3-D} geometry mappings and elastic deformations, to enhance the adaptability of neural networks' for the unique challenges presented by light polarization in polarimetric imaging tasks. Moreover, future work could explore physically consistent implementations of electromagnetic reciprocity, such as simulating reversed source-detector configurations, to leverage deeper symmetry-driven principles in data augmentation. 
%
Overall, our work emphasizes the importance of developing tailored augmentation strategies in Mueller matrix polarimetry, paving the way for improved deep learning models in the field.

\section*{Acknowledgment}
This work was supported by the Swiss National Science Foundation (SNSF) Sinergia Grant No. CRSII5\_205904, "HORAO - Polarimetric visualization of brain fiber tracts for tumor delineation in neurosurgery." We thank the Translational Research Unit, Institute of Tissue Medicine and Pathology, University of Bern, for their assistance with histology and acknowledge UBELIX, the HPC cluster at the University of Bern (https://www.id.unibe.ch/hpc), for computational resources.

\section*{Proof of Intensity Invariance via Calibration}

Assume that at each pixel \(i\) the measured intensity $\mathbf{B}_i$ is:
\[
    \mathbf{B}_i \;=\;\mathbf{A}_i \,\mathbf{M}_i\,\mathbf{W}_i.
\]
We apply a rigid rotation of the polarimeter about its optical axis, assuming a collinear source-detector configuration. In the polarimetry domain, this rotation is represented by an orthogonal matrix $\mathbf{T}_p$. The rotated calibration matrices are:
\[
\mathbf{A}_i' \;=\;\mathbf{A}_i\,\mathbf{T}_p^{-1},
\quad
\mathbf{W}_i' \;=\;\mathbf{T}_p\,\mathbf{W}_i,
\]
and the rotated Mueller matrix is:
\[
\mathbf{M}_i' \;=\;\mathbf{T}_p\,\mathbf{M}_i\,\mathbf{T}_p^{-1}.
\]
Therefore the intensity after rotation is:
\begin{align}
\mathbf{B}_i' 
&= \mathbf{A}_i'\,\mathbf{M}_i'\,\mathbf{W}_i' \nonumber\\
&= (\mathbf{A}_i\,\mathbf{T}_p^{-1})\,
   (\mathbf{T}_p\,\mathbf{M}_i\,\mathbf{T}_p^{-1})\,
   (\mathbf{T}_p\,\mathbf{W}_i) \nonumber\\
&= \mathbf{A}_i\,\mathbf{M}_i\,\mathbf{W}_i
   \;=\;\mathbf{B}_i.
\label{eq:proof_invariance}
\end{align}
This shows that a simultaneous rotation of both source and detector about the common optical axis is fully captured by the transformed calibration matrices $\mathbf{A}'_i$, $\mathbf{W}'_i$ and the raw intensity $\mathbf{B}_i$ remains invariant.

\bibliographystyle{IEEEtran}
\bibliography{sn-bibliography}

\end{document}